\documentclass[runningheads]{llncs}

\usepackage{eccv}

\usepackage{eccvabbrv}
\usepackage{pdfpages}
\usepackage{graphicx}
\usepackage{booktabs}
\usepackage{enumitem}
\usepackage{paralist}
\usepackage{colortbl}
\usepackage[accsupp]{axessibility}
\usepackage{tikz}
\usepackage{pgfplots}
\pgfplotsset{compat=1.18}
\usepgfplotslibrary{fillbetween}
\usepackage{subcaption}
\usepackage{caption}
\usepackage{pifont}
\usepackage{placeins}
\usepackage{threeparttable}
\usepackage{tabularx}
\usepackage[dvipsnames]{xcolor}
\usepackage[dvipsnames]{xcolor}
\newcommand{\mulsign}{\cdot}
\newcommand{\cmark}{\textcolor{Green}{\ding{51}}}
\newcommand{\xmark}{\textcolor{red}{\ding{55}}}
\newcommand{\lightrule}{\arrayrulecolor{gray!40}\midrule\arrayrulecolor{black}}

\newcommand{\supp}{\emph{Supplementary Material}}
\definecolor{fomoblue}{HTML}{126088}
\newcommand{\drop}[1]{\makebox[0pt][l]{\,{\scriptsize$\downarrow$\kern-0.5pt{\tiny #1}}}}
\usepackage{siunitx}
\usepackage[pagebackref,breaklinks,colorlinks,citecolor=eccvblue]{hyperref}

\usepackage{orcidlink}

\begin{document}

\title{Fear of Memorizing Objects:\texorpdfstring{\\}{}End-to-End Zero-Shot Model-Based Object Detection}
 
\titlerunning{Fear of Memorizing Objects}

\author{Leon Greiner\textsuperscript{*}\inst{1,2}\orcidlink{0009-0003-8555-0932} \and
Wout Boerdijk\textsuperscript{*}\inst{1,2}\orcidlink{0000-0003-0789-5970} \and
Junwen Huang\inst{2}\orcidlink{0000-0002-0504-2395} \and
Benjamin Busam\inst{2}\orcidlink{0000-0002-0620-5774} \and
Rudolph Triebel\inst{1,3}\orcidlink{0000-0002-7975-036X} \and
Maximilian Durner\inst{1,2}\orcidlink{0000-0001-8885-5334}}

\authorrunning{L. Greiner et al.}

\institute{German Aerospace Center (DLR), Wessling, Germany\and
Technical University of Munich (TUM), Munich, Germany\and
Karlsruhe Institute of Technology (KIT), Karlsruhe, Germany}

\maketitle
{\let\thefootnote\relax\footnotetext{\textsuperscript{*}Equal contribution}}

\begin{abstract}
This paper presents \textbf{FOMO} (\textbf{F}ear \textbf{o}f \textbf{M}emorizing \textbf{O}bjects), an end-to-end trainable transformer-based architecture for zero-shot \linebreak model-based object detection. Existing approaches typically follow a proposal-and-match pipeline: target-agnostic mask proposals are generated and then matched to template features in a set of loosely coupled stages, making them susceptible to fragmented proposals in clutter and degraded matching under occlusions. FOMO elegantly replaces this paradigm by injecting object template information early in the detection pipeline: Bidirectional attention between learnable latent tokens, multi-view CAD templates, and query image features produces object-aware image features and conditioned queries which are subsequently decoded into bounding-box and confidence predictions. We train FOMO on\linebreak \textbf{Clutter6D}, a novel large-scale synthetic dataset with over 500k images featuring more than 10k diverse object models rendered in highly cluttered scenes with multiple instances and varying occlusion levels, which addresses the limited object diversity and lack of realistic dense clutter in existing datasets for training zero-shot multi-instance algorithms. In true zero-shot fashion, FOMO achieves 54.2 mAP on BOP-Classic-Core at 0.23 s per image, ranking second overall with 3.8~pp difference to the top method while being more than $7\times$ faster. Throughout comprehensive ablation studies we investigate and verify architectural design choices. 
\keywords{Object-detection \and Zero-shot \and Model-based}
\end{abstract}


\begin{figure}[ht]
\centering
\begin{subfigure}[t]{0.6\linewidth}
    \centering
    \includegraphics[width=\linewidth]{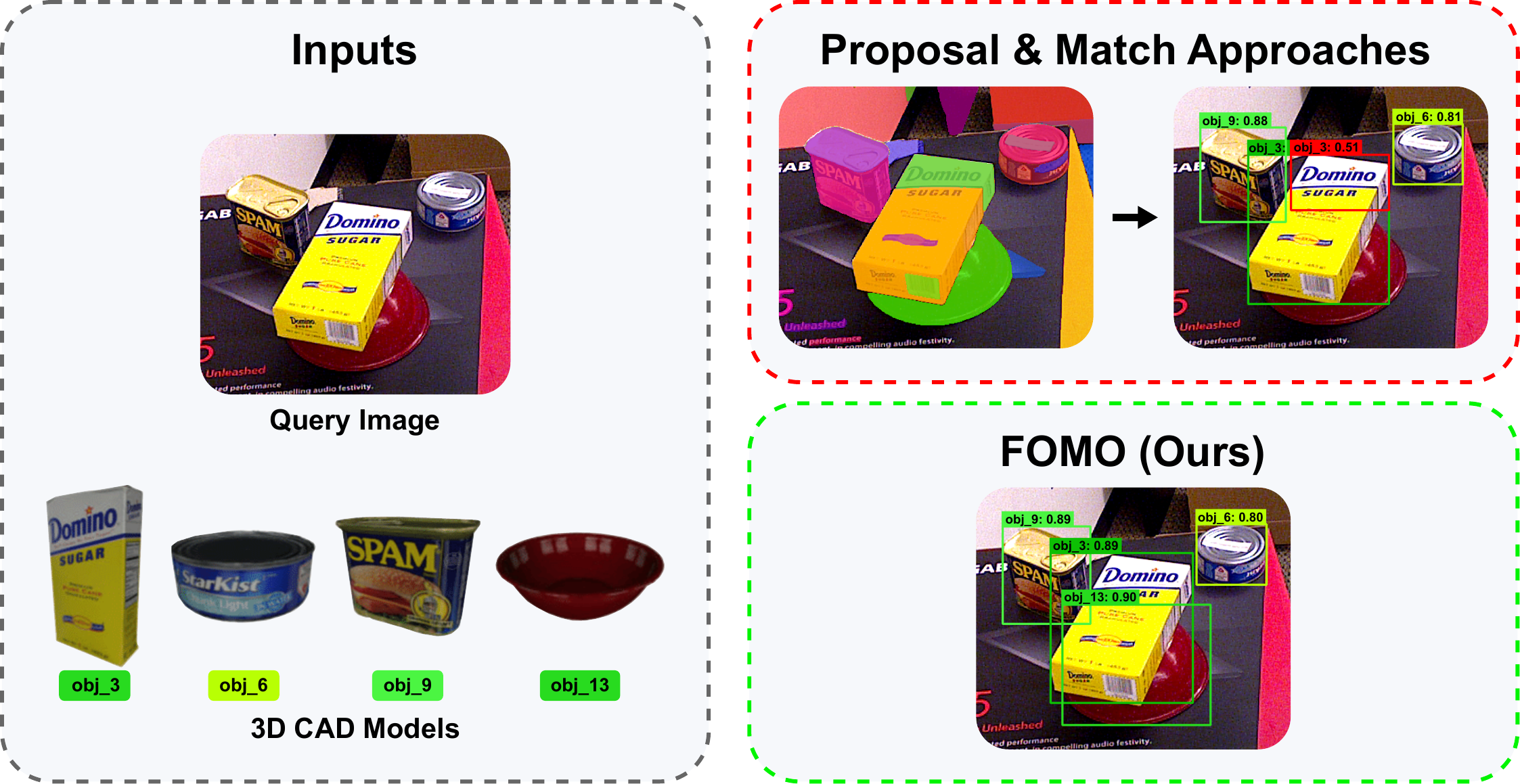}
    \caption{\textbf{Zero-shot object detection setting}}
    \label{fig:fomo_vs_cnos}
\end{subfigure}
\hfill
\begin{subfigure}[t]{0.38\linewidth}
    \begin{minipage}[t]{1\linewidth}
    \centering
    \begin{tikzpicture}
    \begin{axis}[
        width=\linewidth,
        height=0.97\linewidth,
        xlabel={Inference time [s]},
        ylabel={Mean AP [\%]},
        xlabel style={yshift=4pt},
        ylabel style={yshift=-4pt},
        xmode=log,
        xmin=0.15, xmax=10,
        ymin=38, ymax=62,
        xtick={0.2, 0.5, 1, 2, 5},
        xticklabels={0.2, 0.5, 1, 2, 5},
        clip=false,
        grid=both,
        grid style={line width=0.1pt, draw=gray!20},
        major grid style={line width=0.2pt, draw=gray!40},
        legend style={at={(0.03,0.97)}, anchor=north west, font=\tiny, draw=none, fill=white, fill opacity=0.8, text opacity=1, row sep=1pt},
        every axis label/.style={font=\scriptsize},
        tick label style={font=\tiny},
        scatter/classes={
            baseline={mark=*, draw=black, fill=gray!60, mark size=2.0pt}
        },
    ]
    \fill[green!8]
        (axis cs:0.15,62) -- (axis cs:0.15,38)
        -- (axis cs:0.22,38) -- (axis cs:0.22,42.8)
        -- (axis cs:0.25,44.4) -- (axis cs:0.43,50.9)
        -- (axis cs:0.51,53.3) -- (axis cs:1.63,57.9)
        -- (axis cs:7.50,58.0) -- (axis cs:10,58.0)
        -- (axis cs:10,62) -- cycle;
    \draw[green!50!black, thick, dashed, opacity=0.55]
        (axis cs:0.22,38) -- (axis cs:0.22,42.8) -- (axis cs:0.25,44.4)
        -- (axis cs:0.43,50.9) -- (axis cs:0.51,53.3)
        -- (axis cs:1.63,57.9) -- (axis cs:7.50,58.0) -- (axis cs:10,58.0);
    \addplot[scatter, only marks, scatter src=explicit symbolic]
    coordinates {
        (0.22, 42.8) [baseline]
        (0.25, 44.4) [baseline]
        (0.49, 46.9) [baseline]
        (1.06, 47.7) [baseline]
        (0.43, 50.9) [baseline]
        (0.51, 53.3) [baseline]
        (1.63, 57.9) [baseline]
        (7.50, 58.0) [baseline]
    };
    \addplot[only marks, mark=*, mark size=2pt, draw=fomoblue, fill=fomoblue, line width=0.7pt]
    coordinates {(0.23,54.2)};
    
    \node[font=\tiny, anchor=west] at (axis cs:0.22,42.8) {CNOS\scalebox{0.75}{(FastSAM)}};
    \node[font=\tiny, anchor=west] at (axis cs:0.25,44.4) {SAM-6D\scalebox{0.75}{(FastSAM)}};
    \node[font=\tiny, anchor=310] at (axis cs:0.49,46.9) {NIDS-Net};
    \node[font=\tiny, anchor=west] at (axis cs:1.06,47.7) {FOSOD};
    \node[font=\tiny, anchor=west] at (axis cs:0.43,50.9) {F3DT2D};
    \node[font=\tiny, anchor=west] at (axis cs:0.51,53.3) {MUSE};
    \node[font=\tiny, anchor=350] at (axis cs:1.63,57.9) {3PT-Det.\scalebox{0.75}{v2}};
    \node[font=\tiny, anchor=335] at (axis cs:7.50,58.0) {3PT-Det.\scalebox{0.75}{v1}};
    \node[font=\tiny\bfseries, anchor=187, text=fomoblue] at (axis cs:0.23,54.2)  {{\pdfliteral{2 Tr 0.3 w}\textbf{FOMO (Ours)}\pdfliteral{0 Tr 0 w}}};
    \draw[black, line width=0.4pt] (rel axis cs:0,0) rectangle (rel axis cs:1,1);
    \end{axis}
    \end{tikzpicture}
    \caption{\textbf{FOMO Performance}}

    \label{fig:ap_vs_time}
\end{minipage}
\end{subfigure}
\caption{\textbf{FOMO for zero-shot model-based object detection.} 
(a) Conventional zero-shot detectors (\eg, CNOS~\cite{nguyen2023cnos}, top right) require mask proposals (\eg, SAM~\cite{kirillov2023sam}) and match these to object descriptors, leading to duplicate instances (\emph{obj\_3}) or removed predictions due to high occlusions (\emph{obj\_13}). In contrast, FOMO (bottom middle) directly predicts amodal bounding boxes for every prompted object in a single forward pass. 
(b) AP vs.\ inference time on the BOP-Classic-Core unseen-object 2D detection task~\cite{bop_leaderboard_det_unseen}. FOMO achieves the second-highest AP at a fraction of the cost.
}
\end{figure}
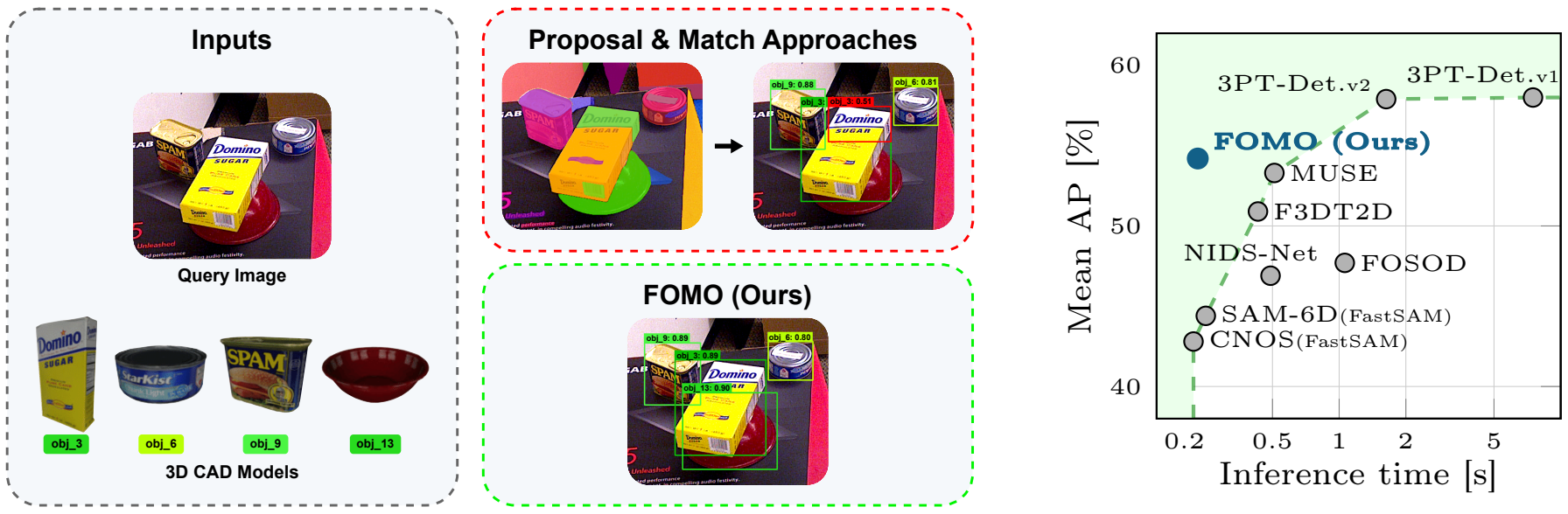
\section{Introduction}
\label{sec:intro}

Accurate object detection is a fundamental computer vision task.
It promotes subsequent tasks such as 6D pose estimation and tracking~\cite{redmon2016yolo, carion2020detr,van2025bop}, and downstream robotic applications such as object grasping and manipulation~\cite{sundermeyer2021contact, newbury2023deep}.

In literature, two contrastive paradigms exist:
On the one end is conventional object detection, where algorithms are trained on annotated data with pre-defined target objects required during training~(\eg,~\cite{ren2016fasterrcnn, carion2020detr, redmon2016yolo, zhu2020deformabledetr}).
On the other end of the spectrum are class-agnostic methods that learn a generic definition of objects~(\eg,~\cite{durner2021unknown, xiang2021learning}).
An integrative link between the two opposing concepts can be described as \emph{prompt-able conditioned detection}:
One stream of literature investigates text-based prompts, also referred to as \emph{open vocabulary detection} (\eg,~\cite{radford2021clip, gu2021vild, minderer2022owlvit, liu2024groundingdino, zang2022ovdetr}).
Yet, text is inherently ambiguous for semantically similar objects (\eg, the prompt ``shoe'' describes an object category and not a specific instance~---~i.e., it does not distinguish between a blue and a red shoe), and many manufactured objects lack canonical names, motivating visual prompts.
Instead, in this work we focus on \emph{model-based zero-shot object detection}, where the CAD model of a target object is only required during inference to generate object-specific features which condition the detection estimation.
The creation of these object descriptors is usually referred to as \emph{onboarding}, where pre-defined viewpoints of the 3D model are rendered to produce object-specific 2D templates.
State-of-the-art approaches~(\eg,~\cite{nguyen2023cnos, cho2025muse, lu2024nidsnet, gandyra2025noctis}) further introduce three subsequent stages~--~\emph{proposal generation}, \emph{embedding}, and \emph{matching}:
First, candidate proposals are generated from a query image, often with Grounding~DINO~\cite{liu2024groundingdino} or SAM~\cite{lin2024sam6d}.
Then, the proposals and the object templates are encoded, usually by a vision foundation model.
Finally, features are matched using a similarity metric to produce predictions for all target objects.

This pipeline has several fundamental limitations: To begin with, proposals are CAD-agnostic, potentially leading to fragmenting or merging instances in heavy clutter.
Second, the discrete proposal--encoding--matching cascade prevents end-to-end learning, and errors introduced early propagate downstream. Furthermore, similarity thresholds are subject to careful tuning and brittle under sim-to-real shifts. Last but not least, mask-based proposals implicitly restrict bounding boxes to a modal format (\emph{i.e.}, reduced to visible object parts).

To overcome these drawbacks we introduce \textbf{FOMO} (\textbf{F}ear \textbf{o}f \textbf{M}emorizing \textbf{O}bjects), to the best of our knowledge the first single-stage, end-to-end trainable detector for model-based zero-shot object detection conditioned on rendered templates. FOMO elegantly redefines proposal matching with a unified architecture that injects template information into the image representation \emph{before} decoding boxes, alleviating the aforementioned restrictions~(\cref{fig:fomo_vs_cnos}). At its core is a novel \textbf{ICMT} (\textbf{I}nterleaved \textbf{C}ross-\textbf{M}odal \textbf{T}ransformer): 
learnable latent tokens alternates between aggregating information from multi-view templates and the query image, and conditioning image features to amplify object-relevant regions. 

To enable end-to-end training at scale, we introduce \textbf{Clutter6D}, a large-scale synthetic dataset with $>$500k images and $>$10k diverse CAD models rendered in highly cluttered, multi-instance scenes with controlled occlusion and distractors. This offers a level of diversity, clutter, and scale not available in existing datasets.

On BOP-Classic-Core, FOMO achieves \textbf{54.2 mean AP} at \textbf{0.23\,s} per query image, outperforming strong training-free baselines~(\cref{fig:ap_vs_time}). 

In summary, the core contributions are: 
(i)~FOMO, a novel single-stage architecture for model-based zero-shot detection that avoids frozen proposal-and-match pipelines;
(ii)~competitive performance on the BOP Benchmark while being more than $7\times$ faster than the top-ranked method;
(iii)~Clutter6D, a large synthetic dataset with $>$10k objects and dense multi-instance clutter designed to train model-based perception algorithms;
and (iv)~detailed ablations on architectural design choices, zero-shot generalization, CAD model influence and~more.


\section{Related Work}
\label{sec:related}


\subsection{Object Detection}

\subsubsection{Closed-Set Detection.}
Modern detectors localize and classify objects from a fixed label set defined at training time and typically cannot generalize to unseen categories at inference~(\eg, \cite{ren2016fasterrcnn, redmon2016yolo}). 
DETR~\cite{carion2020detr} introduces the attention mechanism to the detection task and reframes it as set prediction, removing anchors and NMS via bipartite matching. 
Deformable DETR~\cite{zhu2020deformabledetr} improves convergence and efficiency via sparse learned sampling over multi-scale features.
Our detector builds on this set-prediction paradigm while relaxing the closed-set assumption.

\subsubsection{Open-Vocabulary Detection.}
Open-vocabulary detectors condition detection on natural language prompts, leveraging vision--language pretraining to localize categories specified by text at inference without class-specific retraining. Representative approaches distill CLIP~\cite{radford2021clip} into detectors~\cite{gu2021vild}, attach detection heads to VLM tokens~\cite{minderer2022owlvit}, or fuse language with transformer pipelines~\cite{zang2022ovdetr,liu2024groundingdino}. 

\subsubsection{Zero-Shot Model-Based Detection.}
Zero-shot model-based methods utilize rendered CAD template views as object prompts. 
The state-of-the-art paradigm is a two-stage, training-free pipeline, where proposals of a vision foundation model (\eg,~\cite{kirillov2023sam, zhao2023fastsam, ren2024groundedsam}) 
are compared to the rendered templates of a frozen feature extractor.
In~\cite{chen2024zeropose}, this scheme is leveraged by generating object-agnostic SAM proposals from uniformly sampled point prompts.
Proposal regions are cropped, and their DINOv2 image embeddings are matched against embeddings of offline-rendered CAD templates using cosine similarity, assigning object identities from the highest-scoring template match. 
CNOS~\cite{nguyen2023cnos} subsequently distills this recipe into a seminal template-conditioned detection/segmentation baseline, scoring proposals via DINOv2 CLS-token similarities, and a top-$k$ mean aggregation. 
Subsequent works strengthen different parts of this proposal-and-match paradigm by improving proposal generation and embedding refinement~\cite{lu2024nidsnet}, reformulating the matching step with a joint similarity metric and an uncertainty-aware prior~\cite{cho2025muse}, or general robustness improvements~\cite{gandyra2025noctis}.
In contrast to these two-stage approaches, FOMO seamlessly combines them into a single, end-to-end trainable architecture, focusing on early integration of template information into the model while alleviating hand-crafted similarity metrics.

\subsection{Model-Based Datasets}
\label{sec:related:datasets}

Training model-based zero-shot detectors requires datasets that pair scene images with 3D object models. Real-world benchmarks in the BOP ecosystem~\cite{hodan2024bopchallenge2023detection} capture realistic clutter and occlusion, but contain only a few dozen objects each, 
which is insufficient to learn a generalizable template--conditioned object detection from scratch. Large-scale synthetic datasets~(\eg,~\cite{wang2019normalizedobjectcoordinatespace,he2022fs6d,labbe2022megapose, wen2024foundationpose, zhang2024omni6dpose, zhang2025omni6dlargevocabulary3dobject}) 
offer greater object diversity (up to 42k models), but are constructed for 6D pose estimation from \emph{object-centric training crops}: supervision is centered on a single target instance per crop. Hence, these datasets lack scenes with multiple instances of the same object class and dense image-wide clutter~--~both of which are crucial for training a zero-shot, model-based object detector from scratch. FAT~\cite{tremblay2018falling} addresses multi-instance rendering, but covers only 21 YCB objects, lacking the scale and clutter density required for robust generalization. 
We fill this gap by introducing Clutter6D, a large-scale synthetic dataset with $>$10k diverse 3D object models, dense multi-instance clutter, and heavy occlusions.

\section{Method}
\label{sec:method}

\begin{figure}[tb]
\centering
\includegraphics[width=1.\linewidth]{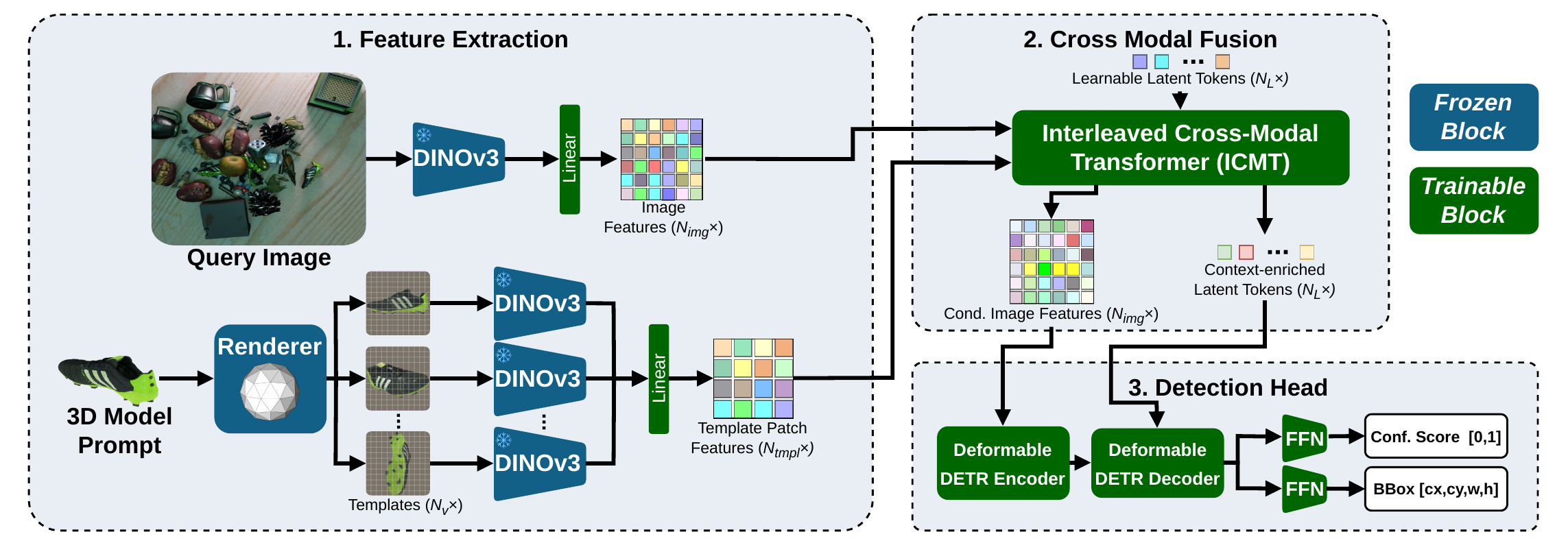}
\caption{\textbf{FOMO architecture.} Rendered template views and the query image are encoded by a shared foundational backbone. The ICMT fuses both modalities, and a detection head produces final predictions. For better visualization only a single object prompt is shown; multi-object prompts are simply batched.
}
\label{fig:architecture}
\end{figure}

We continue by explaining the architecture of FOMO, shown in~\cref{fig:architecture}.
At its~heart is the Interleaved Cross-Modal Transformer (ICMT), which fuses template and image information through learnable latent tokens. 
The ICMT operates on features from a frozen foundational encoder.
It creates conditioned image features and context-enriched object-specific latents, which are subsequently decoded into bounding boxes and confidence scores.
FOMO runs one forward pass per prompted object, reducing detection to binary matchability prediction rather than multi-class classification. These per-object passes are fully independent of one another, with a single exception of the inter-prompt communication layer in the ICMT, which lets each object's tokens briefly interact with those of co-prompted objects on the same image. As a result, the full pipeline roughly scales linearly with the number of prompted objects per image.
In the following, we explain the feature extraction, the ICMT, and the detection head in greater detail.

\begin{figure}[tb]
\centering
\includegraphics[width=1.\linewidth]{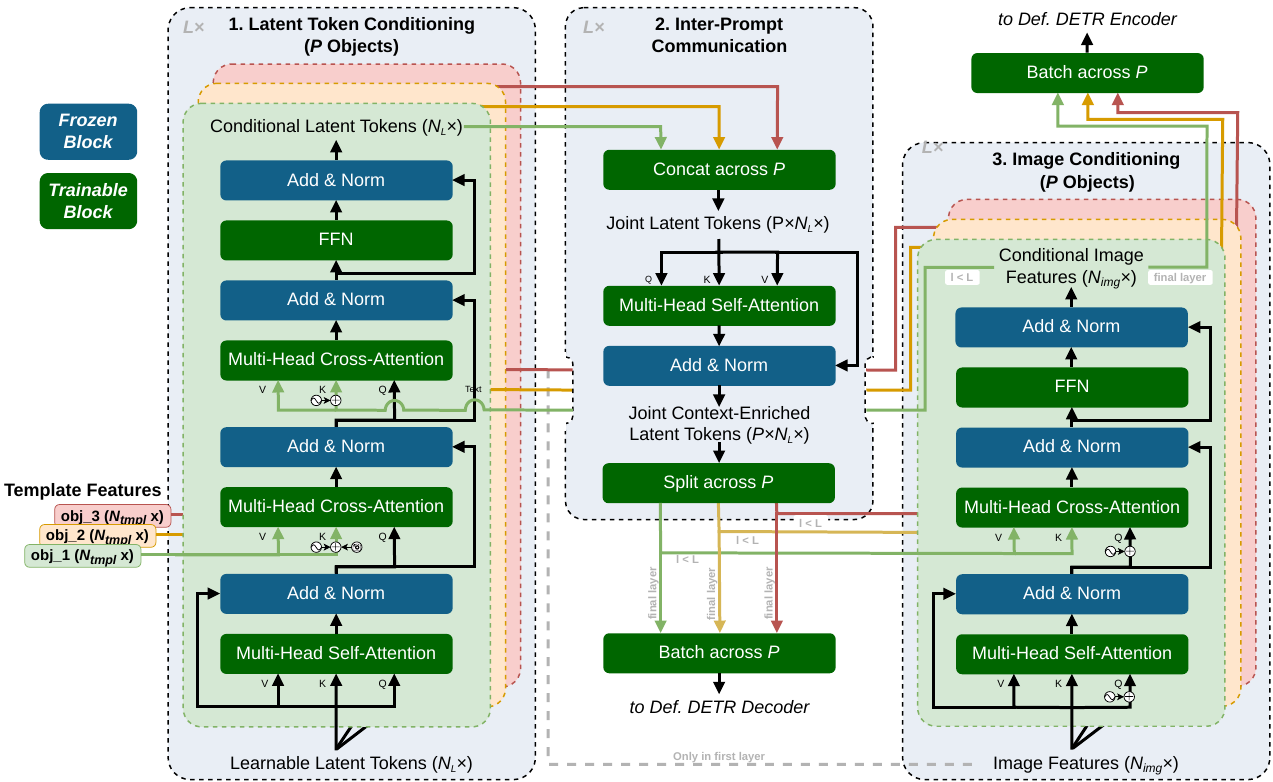}
\caption{\textbf{ICMT detail for multi-object settings.} Each of $L$ layers alternates latent token updates (attending to templates and image) with image feature conditioning via cross-attention to the latent tokens. 
}
\label{fig:icmt_detail}
\end{figure}

\subsection{Feature Extraction}
\label{sec:feature_extraction}

\subsubsection{Template Encoding.}
The 3D CAD model is rendered into $N_v$ views from viewpoints sampled via a Fibonacci lattice~\cite{gonzalez2010measurement}, providing uniform coverage of the object geometry. Each template (resolution $H_{tmpl}\times {W_{tmpl}}$) is processed by a frozen DINOv3 ViT-B/16 backbone, producing $N_{tmpl}{=}N_v\mulsign (H_{tmpl}\mulsign W_{tmpl})/{16^2}$ patch tokens across all views. To preserve spatial and viewpoint structure, each patch is augmented with 2D sinusoidal positional encoding~\cite{vaswani2017attention} (computed per patch grid position) and a learned viewpoint embedding (unique per view). Following DETR~\cite{carion2020detr}, these encodings are added only to the keys $K$\footnote{\emph{NB:} DETR additionally adds encodings to the queries $Q$, yet template features are only utilized as keys $K$ and values $V$ in the ICMT.} in subsequent attention operations, keeping the values $V$ free of positional bias.

\subsubsection{Query Image Encoding.}
The query image of resolution $H_{img}\times {W_{img}}$ is processed by the same DINOv3 backbone, resulting in $N_{img}{=}(H_{img}\mulsign W_{img})/16^2$ patch features. The respective queries $Q$ and keys $K$ are augmented with 2D sinusoidal positional encoding only (no viewpoint embedding). 

\subsubsection{Feature Projection.}
After extraction, both template and query features are passed through separate learnable linear projections to enable task-specific transformations from the frozen backbone dimension.

\subsection{Interleaved Cross-Modal Transformer}
\label{sec:icmt}

At the core of FOMO, the ICMT (\cref{fig:icmt_detail}) compresses the dense template features into a compact set of object-aware representations and simultaneously conditions the query image features on object-specific information. A naive approach would be to directly cross-attend from all $N_{img}$ image patch features to all $N_{tmpl}$ template patches, requiring $\mathcal{O}(N_{img} \mulsign N_{tmpl})$ attention operations per layer. Instead, $N_L$ learnable latent tokens (with $N_L\ll N_{tmpl}$) serve as a compressed bottleneck: latent tokens alternate between attending to template and image features, whereafter the image features attend back to the latent tokens. This reduces operations to $\mathcal{O}(N_L \mulsign N_{tmpl} + 2 \mulsign (N_L \mulsign N_{img}))$ (self-attention costs, shared by both variants, are omitted). A detailed computational ablation in contrast to a naive setting can be found in the \emph{Supplementary} and results in an approximately eight-fold reduction.
The latent tokens are randomly initialized and optimized during training. The ICMT consists of $L$ stacked layers, each comprising three interleaving stages, which are explained in the following.

\subsubsection{Stage 1: Latent Token Update.}
This stage encodes ``\emph{what to look for}'': the canonical object appearance is aggregated into the latent tokens (denoted here as $Z^{(p)}$ for the $p$-th of $P$ object prompts on a given query image) from both the multi-view templates $F_{tmpl}^{(p)}$ and the query image $F_{img}^{(p)}$. The tokens first undergo self-attention, then cross-attend to the templates (collecting the target object information), and finally cross-attend to the query image (grounding this representation in the current scene):
\begin{align}
    Z^{(p)} &\leftarrow \mathrm{SelfAttn}(Q{=}K{=}V{=}Z^{(p)}),\\
    Z^{(p)} &\leftarrow \mathrm{CrossAttn}(Q{=}Z^{(p)},\; K{=}F_{tmpl}^{(p)}{+}\mathrm{PE}_{2D}{+}\mathrm{PE}_{{view}},\; V{=}F_{tmpl}^{(p)}),~\text{and}\\
    Z^{(p)} &\leftarrow \mathrm{CrossAttn}(Q{=}Z^{(p)},\; K{=}F_{img}^{(p)}{+}\mathrm{PE}_{{2D}},\; V{=}F_{img}^{(p)}).
\end{align}
A feed-forward network (FFN) with GELU activation completes the stage. After this, the latent tokens encode a compact description of the target object's visual identity.
As visualized in~\cref{fig:icmt_detail} (left), updating of the latent tokens happens independently of each other.

\subsubsection{Stage 2: Inter-Prompt Communication.}
Here the ICMT encodes ``\emph{who else is looking}'': each object's latent tokens attend to those of all co-prompted objects on the same query image, allowing representations to mutually differentiate and build context about what else is being searched for in the same image. This is achieved by a simple self-attention between the concatenated latent tokens:
\begin{equation}
    Z^{\{1,~\ldots,~P\}} \leftarrow \mathrm{SelfAttn}(Q{=}K{=}V{=}\mathrm{Cat}(Z^{(1)},~\ldots,~Z^{(P)})).
\end{equation}

This communication between different object prompts further boosts the performance of FOMO, as we show in~\cref{sec:ablations:architecture}.

\subsubsection{Stage 3: Image Feature Update.}
This stage encodes ``\emph{where to look}'': it localizes the target by conditioning the image features on the object descriptor latent tokens ($Z$). The image features undergo self-attention, then cross-attend to the latent tokens, allowing each image patch to query whether it corresponds to the target object:
\begin{align}
    F_{img}^{(p)} &\leftarrow \mathrm{SelfAttn}(Q{=}F_{img}^{(p)}{+}\mathrm{PE}_{{2D}},\; K{=}F_{img}^{(p)}{+}\mathrm{PE}_{{2D}},\; V{=}F_{img}^{(p)}),~\text{and}\\
    F_{img}^{(p)} &\leftarrow \mathrm{CrossAttn}(Q{=}F_{img}^{(p)}{+}\mathrm{PE}_{2D},\; K{=}Z^{(p)},\; V{=}Z^{(p)}).
\end{align}
A final FFN produces the conditioned image features. Now, regions containing the target object are amplified while irrelevant background is suppressed. These conditioned image features are used as key-value pairs for the second cross-attention in Stage 1 except for the very first layer, where no updating has occurred yet and the raw backbone features are used instead (gray line in~\cref{fig:icmt_detail}). By alternating these stages across $L$ layers, the ICMT establishes a feedback loop: the latent tokens progressively specialize toward the target object by attending to both its canonical appearance and its manifestation in the scene, while the image features are increasingly primed to emphasize likely object locations. We verify this hypothesis in ablative experiments in~\cref{sec:ablations:icmt}.

\subsection{Detection Head}
\label{sec:detection_head}

The detection head employs the full Deformable DETR~\cite{zhu2020deformabledetr} transformer. The encoder applies deformable self-attention to the conditioned image features from the ICMT, further refining spatial representations. The subsequent decoder uses the context-enriched latent tokens $Z$ as initial object queries, attending to encoder outputs via deformable cross-attention. Since these queries already encode template and scene information, they initialize decoding with strong priors on potential target locations. The decoder outputs feed two prediction heads: (i) a \textit{matchability head} (linear classifier) predicting a confidence score $\hat{p} \in [0,1]$ indicating whether each query corresponds to a target instance, and (ii) a \textit{bounding box head} (3-layer MLP) regressing normalized coordinates $[c_x, c_y, w, h]$.

\subsection{Training Objective}
\label{sec:training_objective}

Following DETR~\cite{carion2020detr}, we formulate detection as set prediction with Hungarian matching~\cite{kuhn1955hungarian}. Since each forward pass is conditioned on a single CAD model, classification reduces to binary matchability (match vs.\ no match), as in OV-DETR~\cite{zang2022ovdetr}. We adopt the Deformable DETR~\cite{zhu2020deformabledetr} matching cost that combines a focal-loss~\cite{lin2017focal} classification term, an $\ell_1$ box distance, and a GIoU cost~\cite{rezatofighi2019generalized}.
For the loss we adopt the Deformable DETR training objective:
\begin{equation}
\label{eq:loss}
\mathcal{L}
= \lambda_{cls} \mathcal{L}_{{cls}}
+ \lambda_{{L1}} \mathcal{L}_{{L1}}
+ \lambda_{{giou}} \mathcal{L}_{{giou}}
+ \lambda_{{embed}} \mathcal{L}_{{embed}},
\end{equation}
where $\mathcal{L}_{{cls}}$ is sigmoid focal loss~\cite{lin2017focal} for matchability, and $\mathcal{L}_{{L1}}$ and $\mathcal{L}_{{giou}}$~\cite{rezatofighi2019generalized} supervise bounding-box regression. Following OV-DETR~\cite{zang2022ovdetr}, we additionally use an embedding reconstruction loss $\mathcal{L}_{{embed}}$, computed as MSE between L2-normalized predicted and conditional query embeddings. This encourages each decoder output to preserve the CAD conditioning by reconstructing its corresponding conditional query embedding (see~\cref{sec:ablations:architecture}).

\section{Clutter6D Dataset}
\label{sec:clutter6d}

Training an end-to-end template-conditioned detector requires large-scale data combining diverse 3D models with cluttered multi-instance scenes. As discussed in \cref{sec:related:datasets}, no existing dataset satisfies these requirements. In order to enable training of FOMO we therefore introduce Clutter6D, a large-scale synthetic dataset providing ground-truth 2D bounding boxes, instance segmentation masks, and 6D object poses, rendered using BlenderProc~\cite{denninger2023blenderproc2}. While FOMO only requires 2D bounding-box annotations for training, the full annotation set enables broader use for other model-based perception tasks. In the following, we describe the dataset in greater detail; for additional information we refer the reader to the \supp.

The dataset uses 10,481 CAD models sourced from GSO~\cite{downs2022gso} (935), OmniObject3D~\cite{wu2023omniobject3d} (5,671), and Objaverse~\cite{deitke2023objaverse} (3,875), quality-filtered for mesh integrity and texture fidelity, spanning 681 categories.

\begin{figure}[tb]
\centering
\begin{subfigure}[t]{0.48\linewidth}
    \centering
    \includegraphics[width=0.49\linewidth]{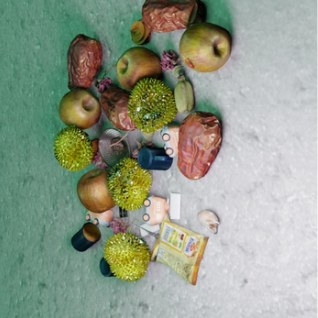}\hfill
    \includegraphics[width=0.49\linewidth]{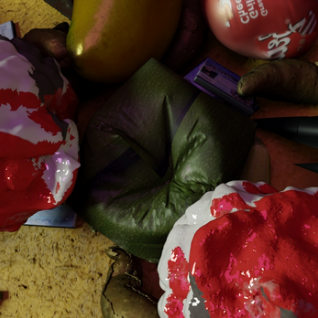}
\end{subfigure}
\hfill
\begin{subfigure}[t]{0.48\linewidth}
    \centering
    \includegraphics[width=0.49\linewidth]{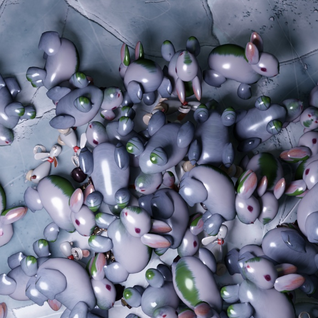}\hfill
    \includegraphics[width=0.49\linewidth]{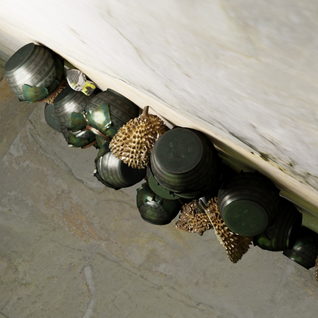}
\end{subfigure}
\caption{\textbf{Clutter6D samples.} \emph{Left:}~Clutter6D-HLD contains many object classes with moderate per-class instance counts. \emph{Right}:~Clutter6D-IND is comprised of fewer classes but many instances per class, yielding heavier same-class clutter and occlusion.}
\label{fig:clutter6d_samples}
\end{figure}

Two subsets are generated: Clutter6D-Household (HLD) features 10--20 object classes per scene with fewer instances each, while Clutter6D-Industrial (IND) uses 2--5 classes with many instances per class. 
In total, 500k images are rendered at $720{\times}720$ resolution across 20k scenes (25 views each), split evenly between HLD and IND (250k images each), with an average of 34.9 instances per image visible~--~significantly denser than existing benchmarks (see \cref{tab:dataset_gap}), with qualitative examples shown in \cref{fig:clutter6d_samples}. 

\begin{table}[tb]
\centering
\caption{\textbf{Clutter6D vs. other model-based synthetic RGB datasets.} Unlike prior large-scale (\emph{i.e.}, \#objects) datasets, Clutter6D supports native multi-instances in every scene and achieves the highest number of objects per image visible, indicating significantly denser, more cluttered, and highly occluded scenes. $^\ast$Inst./Img.~computed based on released annotations; $^\dagger$not reported in associated publication.}
\label{tab:dataset_gap}
\scriptsize
\setlength{\tabcolsep}{2pt}
\begin{tabular*}{\linewidth}{@{\extracolsep{\fill}}lcccccc}
\toprule
Dataset & Objects & Images & Categories & Inst./Img. & Multi-cls. & Multi-inst. \\
\midrule
NOCS CAMERA~\cite{wang2019normalizedobjectcoordinatespace} & 1,085 & 300k & 6 & 4.0$^\ast$ & \cmark & \xmark \\
ShapeNet6D~\cite{he2022fs6d} & 12,490 & 800k & 51 & 5.5$^\ast$ & \cmark & \xmark \\
MegaPose~\cite{labbe2022megapose} & 21,000 & 2M & n/a$^\dagger$ & 13.2$^\ast$ & \cmark & \xmark \\
Omni6D-xl~\cite{zhang2025omni6dlargevocabulary3dobject} & 15,957 & 1.1M & 419 & 6.3$^\ast$ & \cmark & \xmark \\
FoundationPose~\cite{wen2024foundationpose, nvidia_ngc_foundationpose_1_0_1_onnx} & 42,000 & 1.2M & 1,173 & 18.6$^\ast$ & \cmark & \xmark \\
Omni6DPose SOPE~\cite{zhang2024omni6dpose} & 4,162 & 475k & 149 & 15.2$^\ast$ & \cmark & \xmark \\
\midrule
\textbf{Clutter6D (Ours)} & 10,481 & 500k & 681 & 34.9 & \cmark & \cmark \\
\bottomrule
\end{tabular*}
\end{table}

The dataset is split 90/10 into train and validation sets with strict separation of query images, scenes, object classes, and categories.
The training split contains 450,000 images, including 9,531 models from 613 categories, while the validation split contains 50,000 images, including 950 models from 68 categories. This ensures zero overlap in images and object identity between splits.

\section{Experiments}
\label{sec:experiments}

\subsection{Implementation Details and Setup}

\subsubsection{Hyper Parameters and Optimization.}
We set the number of template views to $N_v=12$ ($224{\times}224$ pixels), and keep the resolution of the query images unchanged.
The ICMT uses $L{=}3$ layers with $N_L{=}100$ learnable latent tokens. The Deformable DETR head uses 6 encoder and 6 decoder layers.  Loss weights (cf. \cref{eq:loss}) are set to $\lambda_{cls}{=}\lambda_{L1}{=}5$, $\lambda_{giou}{=}2$, and $\lambda_{embed}{=}1$. We train on Clutter6D with AdamW~\cite{loshchilov2017decoupled} for 10 epochs using cosine annealing.
Each query image is processed with up to 6 positive and 6 negative object prompts~(1:1~ratio), with negatives randomly drawn via hard intra-category sampling (same category, different object), yielding up to 192 training samples per gradient update per GPU. Additional detailed settings are listed in the \supp.

\subsubsection{Data Augmentation.}
We use a two-tier augmentation pipeline applied to both query images and rendered templates: standard photometric transforms (\eg, color jitter, brightness/contrast, blur), and shape-biased augmentations (\eg, grayscale conversion, edge-only rendering, channel shuffling) to suppress texture cues and encourage geometry-driven matching. We schedule the probability of shape-biased augmentations with a cosine annealing policy: strong augmentations early in training to establish robust shape correspondence, then gradually decay to base rates to allow refinement with appearance cues. This reduces reliance on brittle, texture-dependent features while retaining a strong geometric prior (see \cref{sec:ablations:architecture}). For qualitative examples we refer to the \emph{Supplementary}.

\subsubsection{Real-Data Fine-Tuning.}
To expose the model to real-world image statistics while preserving zero-shot generalization, we fine-tune on a mixture of synthetic and real data. Each training batch is sampled from Clutter6D with 70\% probability and from one of seven real datasets with 30\% probability: HOPE~\cite{tyree2022hope}, RU-APC~\cite{rennie2016ruapc}, T-YOL~\cite{hodan2018bop}, XYZ-IBD~\cite{xyz_ibd}, HOT3D~\cite{hot3d}, HANDAL~\cite{handal}, and IPD~\cite{ipd}. Crucially, these datasets are disjoint from the BOP-Classic-Core evaluation benchmark in both objects and scenes, so our model remains \emph{fully zero-shot} at inference on all reported results. The seven real datasets together contain only 161 object classes; retaining Clutter6D in the mix provides substantially higher object diversity and prevents the model from losing its zero-shot capabilities by overfitting to the limited real-data vocabulary.


\subsubsection{Evaluation.}
We evaluate on the seven BOP-Classic-Core~\cite{hodan2018bop} datasets: LM-O~\cite{brachmann2014learning6d}, T-LESS~\cite{hodan2017tless}, TUD-L~\cite{hodan2018bop}, IC-BIN~\cite{doumanoglou2016icbin}, ITODD~\cite{drost2017itodd}, HB~\cite{kaskman2019homebreweddb}, and YCB-V~\cite{xiang2017ycbv}. 
All evaluations are zero-shot: 
None of these objects or scenes appear during training. For component ablations, we additionally report results on the Clutter6D validation set (50k images, 950 objects), which is also zero-shot with respect to the training split due to strict object and scene disjointness (see \cref{sec:clutter6d}). For metrics we follow the standard COCO evaluation protocol~\cite{lin2014microsoft}, reporting Average Precision (AP) averaged over IoU thresholds from 0.50 to 0.95 in steps of 0.05, as adopted by the BOP benchmark~\cite{hodan2018bop}. Further details are listed in the \emph{Supplement}.

\subsection{Results on the BOP-Classic-Core Benchmark}
\label{sec:eval_bop}

The performance of FOMO is demonstrated on the well-established BOP-Classic-Core Benchmark on the task of model-based 2D detection of unseen objects from RGB images only~\cite{bop_leaderboard_det_unseen}.
During inference, an RGB image showing an arbitrary number of instances of a variable number of test objects belonging to a specific dataset, and the respective CAD models of this dataset are provided.

We list results alongside the performance of competing methods in~\cref{tab:bop_results}.
FOMO achieves state-of-the-art results on YCB-V and TUD-L, and ranks second on IC-BIN and HB.
It achieves the second-best performance across all benchmark datasets with a difference of 3.8 percentage points to 3PT-Det~\cite{Kalra_2026_CVPR}, while being $7\times$ faster.
Nonetheless, the model tends to have notable difficulties on T-LESS and ITODD, which we attribute to similar instances (T-LESS) and texture-less objects (both):
Additional analysis of mesh quality is provided in~\cref{sec:ablations:cad_model}.
We discuss further results on T-LESS in the \emph{Supplement}.

\begin{table}[ht]
\centering
\caption{\textbf{Zero-shot detection on BOP core datasets (RGB only).} All data from~\cite{bop_leaderboard_det_unseen}. Time measured per query image averaged over all seven datasets; AP values in~[\%]. \textbf{Best} and \underline{second-best} highlighted. $^\dagger$A second submission achieves 57.9\% mean AP at 1.63\,s.}
\label{tab:bop_results}
\scriptsize
\setlength{\tabcolsep}{2pt}
\begin{threeparttable}
\begin{tabular*}{\linewidth}{@{\extracolsep{\fill}}lccccccccc}
\toprule
Method & Mean & LM-O & T-LESS & TUD-L & IC-BIN & ITODD & HB & YCB-V & Time [s] \\
\midrule
ZeroPose~\cite{chen2024zeropose} & 34.1 & 36.7 & 30.0 & 43.1 & 22.8 & 25.0 & 39.8 & 41.6 & 3.82 \\
CNOS (SAM)~\cite{nguyen2023cnos} & 36.1 & 39.5 & 33.0 & 36.8 & 20.7 & 31.3 & 42.3 & 49.0 & 1.85 \\
CNOS (FastSAM)~\cite{nguyen2023cnos} & 42.8 & 43.3 & 39.5 & 53.4 & 22.6 & 32.5 & 51.7 & 56.8 & \textbf{0.22} \\
NIDS-Net~\cite{lu2024nidsnet} & 46.9 & 45.7 & 49.3 & 48.6 & 25.7 & 37.9 & 58.7 & 62.1 & 0.49 \\
SAM-6D (FastSAM)~\cite{lin2024sam6d} & 44.4 & 43.8 & 41.7 & 54.6 & 23.4 & 37.4 & 52.3 & 57.3 & 0.25 \\
FOSOD$^\ast$ & 47.7 & 46.0 & 47.9 & 49.5 & 23.4 & 42.6 & 56.9 & 67.9 & 1.06 \\
F3DT2D$^\ast$ & 50.9 & 50.4 & 48.2 & 57.3 & 28.4 & 48.0 & 57.7 & 66.6 & 0.43 \\
MUSE~\cite{cho2025muse} & 53.3 & \underline{51.1} & \underline{49.4} & 60.1 & 33.7 & \underline{49.0} & 61.6 & \underline{68.4} & 0.51 \\
3PT-Det.\scalebox{0.75}{v1}\cite{Kalra_2026_CVPR}$^\dagger$ & \textbf{58.0} & \textbf{57.6} & \textbf{52.1} & \underline{66.8} & \textbf{44.2} & \textbf{51.5} & \textbf{68.5} & 65.3 & 7.50 \\
\midrule
\textbf{FOMO (Ours)}\tnote{1} & \underline{54.2} & 50.6 & 36.8 & \textbf{68.0} & \underline{43.0} & 44.3 & \underline{65.4} & \textbf{71.2} & \underline{0.23} \\
\bottomrule
\end{tabular*}

\begin{tablenotes}
\tiny
\item[1] Results are evaluated on the official BOP server.
\end{tablenotes}

\end{threeparttable}

\end{table}

\subsection{Ablation Studies}
\label{sec:ablations}

\subsubsection{Architecture and Training Strategy.}
\label{sec:ablations:architecture}
We ablate architectural design choices on the Clutter6D validation set (\cref{tab:clutter6d_ablations}) and remaining architecture and training components on the full BOP benchmark (\cref{tab:bop_ablation}). Data augmentations and inter-prompt communication are evaluated on BOP rather than Clutter6D: augmentations target the sim-to-real domain shift, which is absent in our synthetic validation setup, while inter-prompt communication primarily benefits scenes with many visually similar object prompts, a condition not reflected in the single-prompt Clutter6D evaluation protocol.

On Clutter6D (\cref{tab:clutter6d_ablations}), the ICMT's bidirectional fusion is the most impactful architectural choice: both the latent-to-image cross-attention in Stage~1 and the image conditioning in Stage~3 contribute complementarily, while their combined removal causes the largest ICMT-related degradation. The Deformable DETR detection head is equally critical, particularly its decoder, where the final fusion between context-enriched latent tokens and conditional image features occurs. The embedding reconstruction loss provides consistent regularization, and upgrading from DINOv2 to DINOv3 yields little gains.

On BOP (\cref{tab:bop_ablation}), real-data fine-tuning has the largest impact, followed by the Clutter6D-specific data augmentations, confirming that both are essential for bridging the sim-to-real gap. Removing inter-prompt communication (Stage~2) primarily hurts on T-LESS, where many visually similar objects co-occur.

\begin{table}[tb]
\centering
\caption{\textbf{Architecture \& training ablation on Clutter6D.} AP ([\%]) on the Clutter6D validation set. All variants were trained for 10 epochs on synthetic Clutter6D.} 
\label{tab:clutter6d_ablations}
\scriptsize
\setlength{\tabcolsep}{4pt}
\begin{tabular*}{\linewidth}{l@{\hspace{1.cm}}@{\extracolsep{\fill}}ccc}
\toprule
Configuration & Mean & HLD & IND \\
\midrule
w/o ICMT Stage 3 (late fusion) & 55.4\drop{6.2} & 57.4 & 53.3 \\
w/o latent-to-image cross-attn (w/o Stage 3) & 53.7\drop{7.9} & 55.2 & 52.1 \\
w/o latent-to-image cross-attn (w/ Stage 3) & 58.2\drop{3.4} & 61.5 & 54.8 \\
\lightrule
w/o Deformable DETR encoder & 57.2\drop{4.4} & 61.1 & 53.2 \\
w/o Deformable DETR encoder + decoder & 47.0\drop{14.6} & 51.8 & 42.1 \\
\lightrule
w/o embedding reconstruction loss & 59.7\drop{1.9} & 62.1 & 57.3 \\
\lightrule
DINOv2 backbone (instead of DINOv3) & 59.6\drop{2.0} & 61.6 & 57.5 \\
\midrule
\textbf{Full model} & 61.6 & 64.9 & 58.2\\
\bottomrule
\end{tabular*}
\end{table}

\begin{table}[tb]
\centering
\caption{\textbf{Architecture \& training ablation on BOP.} AP ([\%]) on BOP core datasets. All configurations include 1200 steps of real-data fine-tuning unless stated otherwise.}
\label{tab:bop_ablation}
\scriptsize
\setlength{\tabcolsep}{2pt}
\begin{tabular*}{\linewidth}{l@{\hspace{.5cm}}@{\extracolsep{\fill}}c@{\hspace{14pt}}ccccccc}
\toprule
Configuration & Mean & LM-O & T-LESS & TUD-L & IC-BIN & ITODD & HB & YCB-V \\
\midrule
w/o real-data fine-tuning & 43.7\drop{10.5} & 44.2 & 26.2 & 45.5 & 28.5 & 30.9 & 63.3 & 67.1 \\
w/o Clutter6d data aug. & 48.2\drop{6.0} & 50.5 & 21.5 & 59.7 & 42.0 & 30.9 & 63.3 & 69.6 \\
w/o ICMT Stage 2 & 52.3\drop{1.9} & 51.0 & 28.9 & 72.3 & 40.6 & 38.5 & 63.0 & 72.1 \\
\midrule
\textbf{Full model} & 54.2 & 50.6 & 36.8 & 68.0 & 43.0 & 44.3 & 65.4 & 71.2 \\
\bottomrule
\end{tabular*}
\end{table}

\subsubsection{ICMT Feature Enhancement.}
\label{sec:ablations:icmt}

\begin{figure}[tb]
\centering
\includegraphics[width=\linewidth]{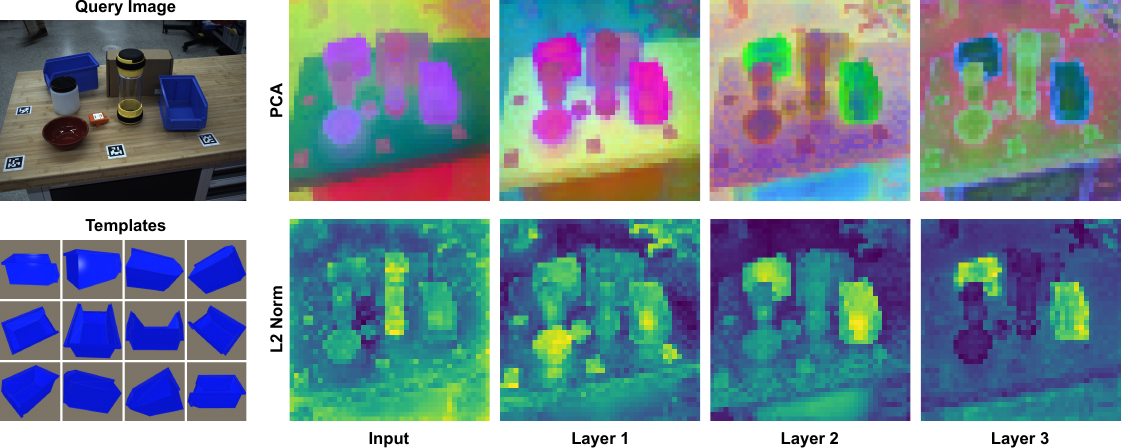}
\caption{\textbf{Query image feature evolution across ICMT layers} on a real-world scene with multiple instances and classes.
\textit{Left:} Query image (top) and 12 template views (bottom).
\textit{Top row:} PCA projection of patch features to RGB, showing progressive semantic clustering around the target.
\textit{Bottom row:} L2 norm of feature vectors, revealing concentration at object locations.
}
\label{fig:icmt_layers}
\end{figure}

A key design goal of the ICMT is to transform generic backbone features into object-specific representations. To verify this behavior, query image features at each ICMT layer are visualized using PCA projection to RGB and L2 norm magnitude maps in~\cref{fig:icmt_layers}. The input features from DINOv3 exhibit diffuse activations without target-object organization. After the first ICMT layer, initial signs of object-aware processing emerge; by layer~3 the L2 norm maps approximate coarse segmentation masks at patch resolution. 
This emergent behavior, without explicit mask supervision, shows that ICMT amplifies features of the prompted object while suppressing background, producing localized features for the decoder.

Analysis of the latent-to-image cross-attention weights in Stage~1 across the full Clutter6D validation set (50k images) reveals progressive attention sharpening: the Gini coefficient increases monotonically from layer~1 to layer~3 (see \supp). Early layers perform broad information gathering, while later layers concentrate on object-relevant patches.

Furthermore, the 100 learnable latent tokens self-organize during training to tile the image space spatially, with each token specializing in a different region, without any explicit supervision (\cref{fig:cad_token_locality}). 
Notably, no semantic or category-based specialization is observed: each active token detects objects from dozens of categories and classes in proportion to their overall frequency rather than specializing in specific object types. This aligns with FOMO's zero-shot design: tokens learn to specialize purely on geometric properties (spatial location and object scale) that transfer across novel categories.

\newlength{\pairfigH}
\setlength{\pairfigH}{4cm}

\begin{figure}[t]
\centering
\begin{minipage}[t]{0.56\linewidth}
\centering
\vspace{0pt} 

\begin{minipage}[c][\pairfigH][c]{\linewidth}
\centering
\includegraphics[width=\linewidth,height=\pairfigH,keepaspectratio]{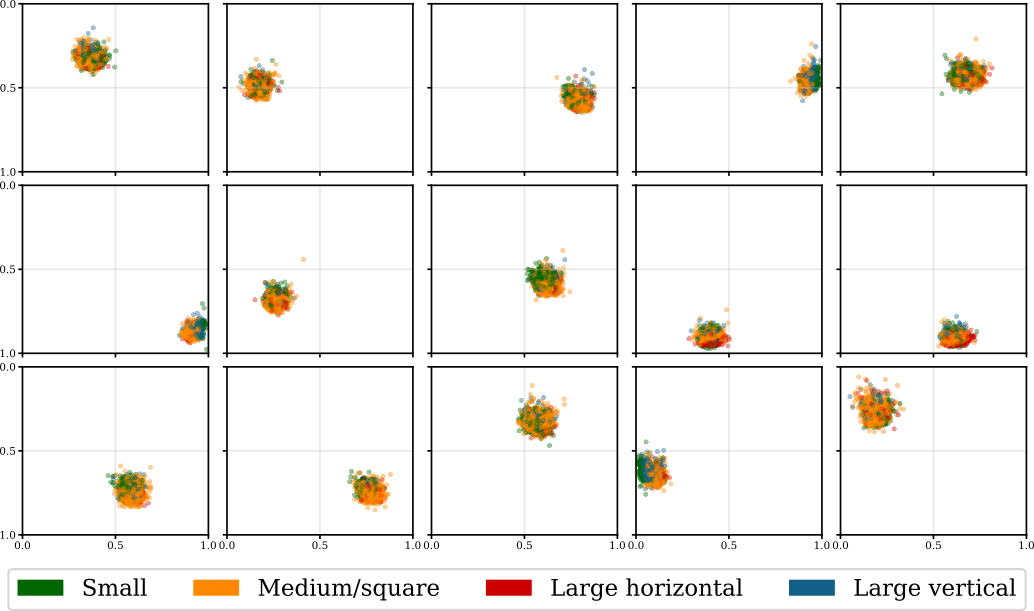}
\end{minipage}

\captionof{figure}{\textbf{Latent token spatial specialization.} Each subplot shows the spatial distribution of detection centers per latent token (15 of 100 shown) on the Clutter6D validation set (50k images, 175k matched predictions). Tokens self-organize to tile the image space without explicit supervision.}
\label{fig:cad_token_locality}
\end{minipage}
\hfill
\begin{minipage}[t]{0.42\linewidth}
\centering
\vspace{0pt}

\begin{minipage}[c][\pairfigH][b]{\linewidth}
\centering
\begin{tikzpicture}
\begin{axis}[
    width=4.1cm,
    height=3.1cm,
    scale only axis,
    xlabel={Epoch},
    ylabel={AP [\%]},
    xlabel style={yshift=3pt, font=\small},
    ylabel style={yshift=-4pt, font=\small},
    xmin=0, xmax=10,
    ymin=0, ymax=70,
    xtick={0,2,...,10},
    ytick={0,10,...,70},
    grid=both,
    grid style={line width=0.1pt, draw=gray!20},
    major grid style={line width=0.2pt, draw=gray!40},
    every axis label/.style={font=\scriptsize},
    tick label style={font=\tiny},
    legend style={
        at={(0.97,0.03)},
        anchor=south east,
        font=\small,
        draw=none,
        fill=white,
        fill opacity=0.85,
        text opacity=1,
        row sep=2pt,
        /tikz/every even column/.append style={column sep=5pt},
    },
    legend cell align={left},
    clip=true,
]
    \addplot[draw=none, forget plot, name path=train]
        table[x=epoch, y=ap, col sep=comma] {figures/train_ap.csv};
    \addplot[draw=none, forget plot, name path=val]
        table[x=epoch, y=ap, col sep=comma] {figures/val_ap.csv};
    \addplot[fill=fomoblue, fill opacity=0.06, forget plot]
        fill between[of=train and val];

    \addplot+[
        smooth,
        mark=none,
        thick,
        color=black!60,
        densely dashed,
    ] table[x=epoch, y=ap, col sep=comma] {figures/train_ap.csv};
    \addlegendentry{Train AP}

    \addplot+[
        mark=*,
        mark size=1.2pt,
        thick,
        color=fomoblue,
        mark options={fill=fomoblue, draw=fomoblue},
    ] table[x=epoch, y=ap, col sep=comma] {figures/val_ap.csv};
    \addlegendentry{Val AP}
\end{axis}
\end{tikzpicture}
\end{minipage}
\captionof{figure}{\textbf{Train vs.\ validation AP across epochs on Clutter6D.} The similar performance on disjoint object sets between both splits indicates true zero-shot generalization rather than object memorization.}
\label{fig:fomo_generalization}
\end{minipage}
\end{figure}

\subsubsection{The Fear of Memorizing Objects.}

A significant premise for zero-shot object detection algorithms is its generalization capability onto novel object classes and instances. \Cref{fig:fomo_generalization} visualizes mean AP on the Clutter6D train and validation set.
The latter, as readily discussed in~\cref{sec:clutter6d}, is solely comprised of objects and query image scenes disjoint from the training set.
This highlights FOMO's potential for true zero-shot generalizability, and further suggests that the reason for the variance in performance on the seven BOP core datasets is likely to be attributed to dataset-dependent properties, \eg, CAD model quality.

\subsubsection{CAD Model Quality.}
\label{sec:ablations:cad_model}

One potential sim-to-real gap originates from different CAD model qualities between training and test environments, hence potentially influencing the performance reported in \cref{tab:bop_results}.
To investigate this, we take a subset of models of the YCB-V dataset, where FOMO performs best within BOP core, and successively degrade both texture and geometry.
To this end, the 4k texture is resized to 1/32th resolution.
For the geometry we perform vertex degradation by remeshing and shrinkwrapping, the latter ensures valid texture UV mapping. 
\cref{tab:ablation:cad_model_quality} gives an overview of AP values across three object super-categories~--~box-like objects, cylindrical objects, and other.
Shape degradation mostly influences objects with distinct shapes (D-S), while it is the other way around for texture degradation (D-T).
There, visual features are insufficient to distinguish between object instances of similar shapes.
Consequently, removing texture completely (No T) diminishes results on these object types.
The most extreme case (D-S+No T) results in the worst performance.
A full table for all 21 YCB-V objects with example qualitative templates as well as further details on the degradation are provided in the~\supp. 

\begin{table}[ht]
\centering
\caption{Mean AP values ([\%]) across YCB-V objects merged into super-categories. \mbox{D-S = degraded} shape; \mbox{D-T~=~degraded} texture.}
\scriptsize

\begin{tabularx}{\linewidth}{l@{\hspace{1.cm}}@{\extracolsep{\fill}}SSSSSS}
\toprule
Object Type & {Baseline} & {D-S} & {D-T} & {D-S+T} & {No T} & {D-S+No T} \\
\midrule
Box & 73.8 & 75.0 & 60.3 & 60.5 & 12.3 & 14.4 \\
Cylindric & 74.1 & 75.1 & 53.6 & 55.3 & 12.6 & 4.7 \\
Misc & 68.4 & 66.2 & 66.5 & 65.1 & 59.6 & 37.6 \\
\midrule
\textbf{Mean across categories} & 72.1 & 72.1 & 60.1 & 60.3 & 28.2 & 18.9 \\
\bottomrule
\end{tabularx}
\label{tab:ablation:cad_model_quality}
\end{table}


\subsection{Limitations}

While FOMO ranks first on two datasets, and performs competitively on additional three, and substantially excels in speed in contrast to competing methods, its performance heavily relies on accurate CAD models (cf. \cref{sec:ablations:cad_model}).
Furthermore, as seen on T-LESS, multiple similar objects (both within the scene and as prompts) pose difficulties: Here, the ICMT is unable to differentiate object features, albeit inter-prompt communications, leading to incorrect predictions. 
Refer to the supplement for a qualitative investigation.

\section{Conclusion}
\label{sec:conclusion}

We presented FOMO, a novel end-to-end single-stage framework for model-based zero-shot object detection.
In true zero-shot fashion, FOMO achieves strong results on challenging real-world benchmark datasets while being substantially faster than competing methods.
Additionally, we conducted rigorous ablation studies on various architectural design choices, and highlighted the contributions of different stages in the ICMT as a vital core element of our method.
FOMO promotes a shift from closed-set detection algorithms toward end-to-end trainable model-conditioned perception. 
This approach opens the path for visual systems, extending model-conditioned inference beyond detection to segmentation, pose estimation, and tracking, ultimately enabling agile embodied systems to adapt to new objects in open-world environments.


%

\bibliographystyle{splncs04}
\bibliography{main,supplementary}

\title{Supplementary Material}

\titlerunning{Supplementary Material}

\author{}

\authorrunning{L. Greiner et al.}

\institute{}

\maketitle

\renewcommand{\thesection}{\Alph{section}}

This Supplementary Material provides complementary details to our paper \textbf{Fear of Memorizing Objects: End-to-End Zero-Shot Model-Based Object Detection}.
We start by listing additional training details (parameters, hardware, runtime, augmentation examples) in~\cref{sec:training_details}.
We then provide evaluation details, including batching strategy and a per-dataset timing ablation in~\cref{sec:eval_details}.
We proceed by providing insights on the ICMT and discuss latent token statistics and computation ablation to a naive variant~(\cref{sec:icmt}).
Next, we explain Clutter6D in greater detail and list rendering settings, template rendering, object model details, and further qualitative examples in~\cref{sec:clutter6d}.
Finally, we provide details on the CAD model degradation ablation study, and visualize features for a T-LESS failure case in~\cref{sec:misc}.

\section{Additional Training Details}
\label{sec:training_details}

\subsection{Model Training and Hyperparameters}

Table~\ref{tab:hyperparams} provides a comprehensive reference of all hyperparameters used for model architecture configuration, optimization, and data loading.

\begin{table}[tb]
	\centering
	\caption{\textbf{Model architecture and training hyperparameters.}}
	\tiny
	\setlength{\tabcolsep}{4pt}
	\begin{tabularx}{0.9\linewidth}{@{}l l X@{}l}
		\toprule
		\textbf{Parameter} & \textbf{Value} & \textbf{Notes} \\
		\midrule
		\multicolumn{3}{@{}l}{\textit{Backbone}} \\
		Architecture & DINOv3 ViT-B/16 & \\
		Hidden dimension & 768 & \\
		\midrule
		\multicolumn{3}{@{}l}{\textit{ICMT}} \\
		Number of layers & 3 & \\
		Attention heads & 12 & \\
		Feed-forward dimension & 3{,}072 & \\
		Dropout & 0.1 & \\
		Latent tokens & 100 & Conditioned Object Queries \\
		\midrule
		\multicolumn{3}{@{}l}{\textit{Deformable DETR}} \\
		Encoder layers & 6 & \\
		Decoder layers & 6 & \\
		Attention heads & 12 & \\
		Feed-forward dimension & 3{,}072 & \\
		Dropout & 0.1 & \\
		Auxiliary loss & \cmark & Applied to all decoder layers \\
		\midrule
		\multicolumn{3}{@{}l}{\textit{Optimizer}} \\
		Type & AdamW & Decoupled weight decay \\
		Learning rate & $5 \cdot 10^{-5}$ & \\
		Minimum learning rate & $10^{-7}$ & Cosine annealing target \\
		Weight decay & $5 \cdot 10^{-3}$ & \\
		Gradient clipping & 0.1 & Max norm \\
		\midrule
		\multicolumn{3}{@{}l}{\textit{Training}} \\
		Epochs & 10 & \\
		Warmup epochs & 1 & Linear warmup \\
		Batch size & 1 & Per GPU \\
		Gradient accumulation & 16 & Effective batch size: 16 \\
		Precision & bfloat16 & Mixed precision \\
		Validation frequency & 0.25 & Every quarter epoch \\
		\midrule
		\multicolumn{3}{@{}l}{\textit{Loss/Matching weights}} \\
		Classification ($\lambda_{\text{cls}}$) & 5 & Focal loss \\
		L1 box ($\lambda_{\text{L1}}$) & 5 & \\
		GIoU ($\lambda_{\text{giou}}$) & 2 & \\
		Embedding ($\lambda_{\text{embed}}$) & 1 & Reconstruction loss \\
		Focal $\alpha$ / $\gamma$ & 0.5 / 2 & \\
		\midrule
		\multicolumn{3}{@{}l}{\textit{Data loading}} \\
		Template views & 12 & Farthest-point sampled from 42 \\
		Template resolution & $224 \times 224$ & \\
		Query resolution & $720 \times 720$ & \\
		Negative-to-positive ratio & 1:1 & \\
		Max positives per sample & 8 & \\
		Min visibility (positive) & 0.3 & \\
		Min visibility (loss) & 0.1 & \\
		\bottomrule
	\end{tabularx}
	\label{tab:hyperparams}
\end{table}

\subsection{Hardware and Runtime}
Training was conducted on 4$\times$NVIDIA A100 80\,GB GPUs and required approximately 224 hours of wall-clock time for the full 10-epoch schedule (4.5\,M query-image forward passes in total) on the Clutter6D training set. The real-data fine-tuning was conducted on a single NVIDIA H200 141\,GB GPU for 1200 gradient steps. All inference timings are measured on a single NVIDIA H200 141\,GB GPU following the BOP protocol~\cite{hodan2018bop}: the reported time covers the complete detection pipeline from the point right after raw data (image, 3D models) is loaded until final predictions are available, excluding offline template rendering and data loading. In practice, query image backbone features are computed once per image and shared across all object prompts, and template features are pre-computed and cached per object during onboarding since they are independent of the query image.

\subsection{Qualitative Augmentation Examples}

\begin{figure}[!ht]
	\centering
	\begin{subfigure}[t]{0.4\textwidth}
		\centering
		\includegraphics[width=\linewidth]{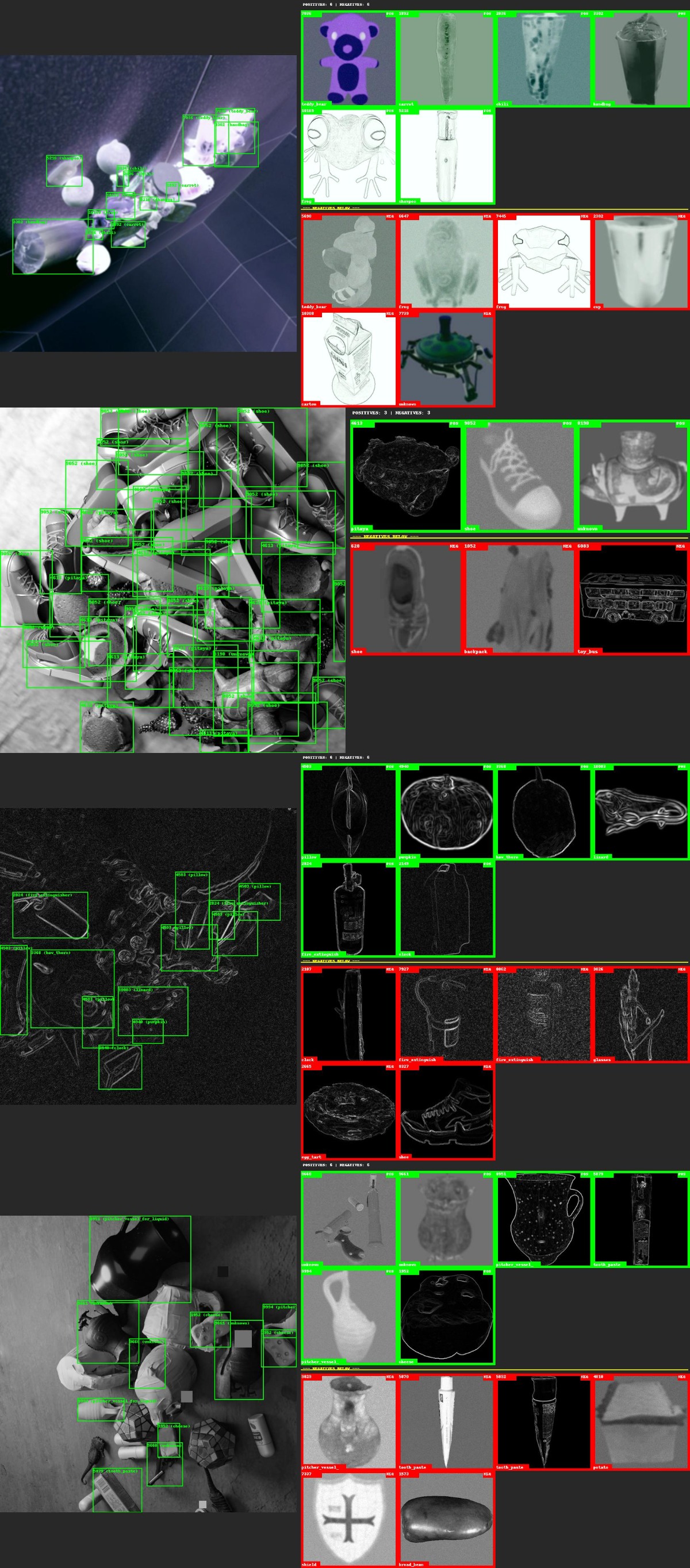}
		\caption{Shape-focused phase (Epoch~0)}
		\label{fig:augs_epoch0}
	\end{subfigure}
	\hspace{1em}
	\begin{subfigure}[t]{0.4\textwidth}
		\centering
		\includegraphics[width=\linewidth]{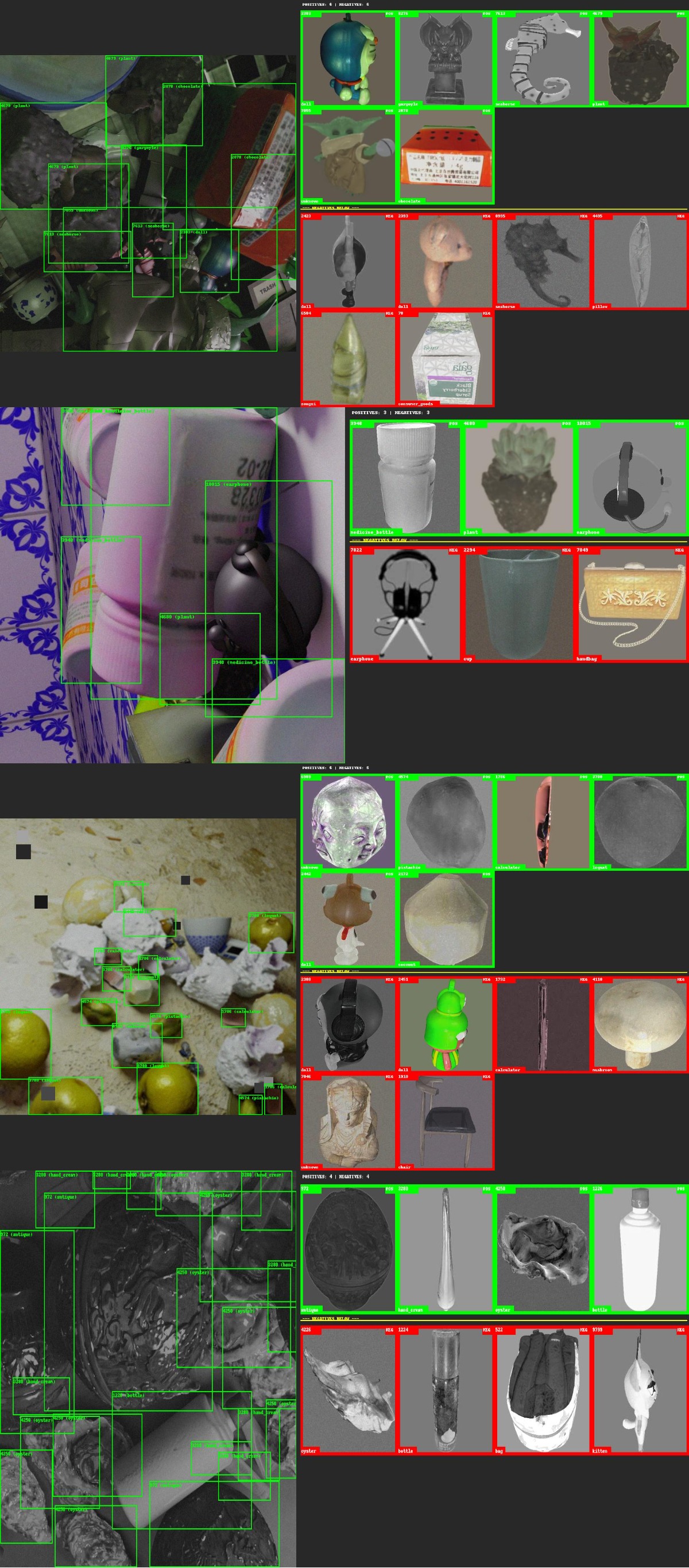}
		\caption{Base augmentation phase (Epoch~6)}
		\label{fig:augs_epoch6}
	\end{subfigure}
	\caption{\textbf{Effect of augmentation scheduling on training samples.} (a)~Shape-focused phase applies aggressive augmentations including high rates of grayscale conversion, edge-only rendering, and texture suppression, forcing the model to learn geometric matching. (b)~Base augmentation phase shows the decayed augmentation rates after the cosine schedule, allowing texture-based refinement while retaining the geometric foundation. In the query images, ground-truth bounding boxes are shown in \textit{\textcolor{green}{green}}. Next to each query, the first template view of all prompted objects is displayed: templates with a \textit{\textcolor{green}{green}} frame indicate positive prompts (objects present in the scene), while templates with a \textit{\textcolor{red}{red}} frame indicate negative prompts (objects absent from the scene).}
	\label{fig:augmentation_samples}
\end{figure}

\Cref{fig:augmentation_samples} visualizes the effect of the augmentation scheduling strategy described in Sec. 5.1, contrasting training samples from the early shape-focused phase with samples from the later base augmentation phase.

\FloatBarrier

\section{Evaluation Details}
\label{sec:eval_details}

\subsection{Inference Strategy}
\label{sec:eval_details:inference}

FOMO runs one forward pass per object prompt (cf.\ Sec.~3 of the main paper), so inference time scales with the number of prompted objects per dataset.
To maximize GPU utilization, we apply two levels of batching:
First, all $P$ object prompts of a dataset are processed jointly for each query image (\emph{object-prompt batching}).
This is not merely an efficiency choice but a functional requirement: the inter-prompt communication layer (Stage~2 of the ICMT) performs self-attention over the concatenated latent tokens of all co-prompted objects, so all prompts must be present in the same forward pass.
Second, multiple query images are batched with a batch size of~4 on the H200 (\emph{query-image batching}), reducing per-image overhead from kernel launches and data transfers.
The resulting effective batch size is $P \times 4$; for example, on YCB-V ($P{=}21$) this yields 84 forward passes processed in parallel.

Two caching strategies further reduce redundant computation:
Template features are pre-computed once per object during onboarding and reused across all query images, as they are independent of the scene; this cost is amortized and excluded from per-image timing.
Similarly, the DINOv3 backbone encodes each query image exactly once, and the resulting patch features are shared across all object prompts on that image, making this encoding cost constant regardless of~$P$.
The per-image cost is therefore dominated by the ICMT and detection-head forward passes, which scale linearly with~$P$.
The inter-prompt communication layer introduces a quadratic term $\mathcal{O}((P \cdot N_L)^2)$ in the number of prompts; however, this overhead remains small relative to the remaining per-prompt stages, so overall inference time scales approximately linearly with $P$ (see \cref{fig:token_analysis_c}).
For datasets with many objects, the marginal per-object cost converges to ${\approx}0.013$\,s on a H200 and ${\approx}0.033$\,s on a RTX~3090, with a constant overhead from backbone encoding that dominates only for few-object datasets.

\subsection{Per-Dataset Timing Ablation}
\label{sec:eval_details:timing}

\Cref{tab:timing_ablation} reports per-image inference times on two GPUs across all seven BOP-Classic-Core datasets: a NVIDIA H200 (141\,GB, used for the official BOP submission), and a NVIDIA RTX~3090 (24\,GB). We additionally report RTX~3090 numbers because the current BOP benchmark does not prescribe a fixed compute platform. For instance, 3PT-Det. reports on an NVIDIA V100, MUSE~\cite{cho2025muse} on an RTX~4090, and others on a Quadro P6000, making direct runtime comparisons across methods difficult~\cite{bop_leaderboard_det_unseen}.
The timings confirm approximately linear scaling with $P$ (see \cref{fig:token_analysis_c}).

\begin{table}[ht]
	\centering
	\caption{\textbf{Per-image inference time [s]} on NVIDIA H200 and RTX~3090 across BOP-Classic-Core datasets. Number of object prompts $P$ per dataset shown in parentheses. Query-image batch size is 4 (H200) and 1 (RTX~3090).}
	\label{tab:timing_ablation}
	\scriptsize
	\setlength{\tabcolsep}{2pt}
	\begin{tabular*}{\linewidth}{@{\extracolsep{\fill}}lcccccccc}
		\toprule
		& & LM-O & T-LESS & TUD-L & IC-BIN & ITODD & HB & YCB-V \\
		GPU & Mean & {\tiny($P{=}8$)} & {\tiny($P{=}30$)} & {\tiny($P{=}3$)} & {\tiny($P{=}2$)} & {\tiny($P{=}28$)} & {\tiny($P{=}33$)} & {\tiny($P{=}21$)} \\
		\midrule
		H200 & 0.233 & 0.117 & 0.375 & 0.060 & 0.047 & 0.353 & 0.410 & 0.269 \\
		RTX 3090 & 0.610 & 0.302 & 0.975 & 0.160 & 0.133 & 0.923 & 1.069 & 0.706 \\
		\bottomrule
	\end{tabular*}
\end{table}


\section{ICMT Details}
\label{sec:icmt}
\subsection{Latent Token Attention Statistics}

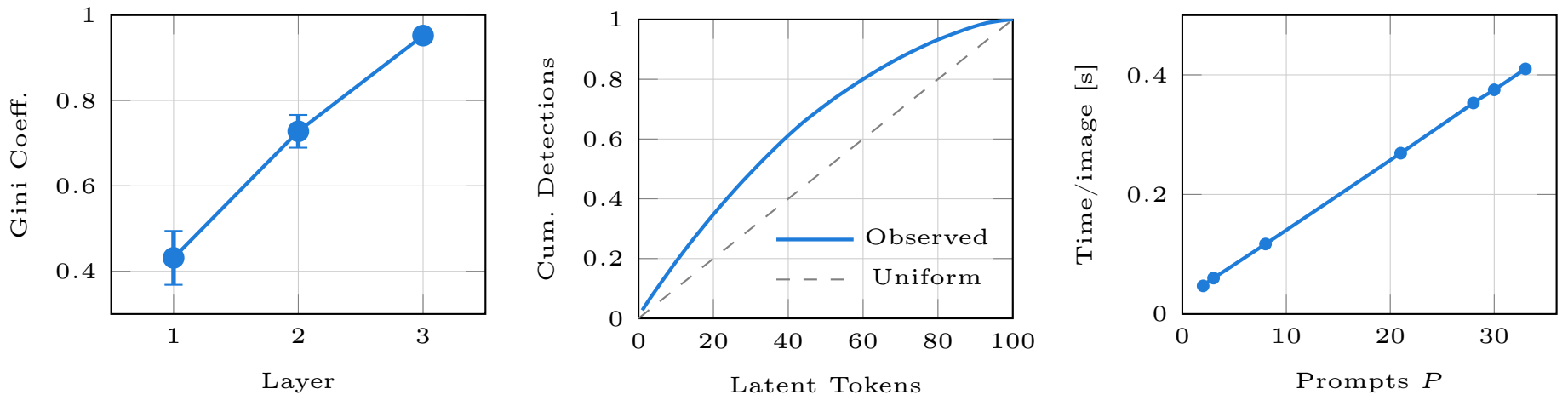
\begin{figure}[tb]
	\centering
	\begin{subfigure}{0.32\linewidth}
		\centering
		\begin{tikzpicture}
			\begin{axis}[
				width=1.15\linewidth,
				height=\linewidth,
				xlabel={Layer},
				ylabel={Gini Coeff.},
				xmin=0.5, xmax=3.5,
				ymin=0.3, ymax=1.0,
				xtick={1,2,3},
				grid=both,
				grid style={line width=0.1pt, draw=gray!20},
				major grid style={line width=0.2pt, draw=gray!40},
				legend pos=south east,
				legend style={font=\tiny},
				tick label style={font=\tiny},
				label style={font=\tiny},
				]
				\addplot+[
				thick,
				mark=*,
				mark options={scale=1.0},
				color=eccvblue,
				error bars/.cd,
				y dir=both,
				y explicit,
				error bar style={line width=1pt, color=eccvblue}
				] table [x=layer, y=gini_mean, y error=gini_std, col sep=comma] {figures/gini_coefficient_layers.csv};
			\end{axis}
		\end{tikzpicture}
		\caption{\textbf{Attention Sparsity}}
		\label{fig:token_analysis_a}
	\end{subfigure}
	\hfill
	\begin{subfigure}{0.32\linewidth}
		\centering
		\begin{tikzpicture}
			\begin{axis}[
				width=1.15\linewidth,
				height=\linewidth,
				xlabel={Latent Tokens},
				ylabel={Cum. Detections},
				xmin=0, xmax=100,
				ymin=0, ymax=1.0,
				grid=both,
				grid style={line width=0.1pt, draw=gray!20},
				major grid style={line width=0.2pt, draw=gray!40},
				legend style={at={(1,0.03)}, anchor=south east, font=\tiny, draw=none, fill=none},
				tick label style={font=\tiny},
				label style={font=\tiny},
				]
				\addplot[
				thick,
				color=eccvblue,
				] table [x=num_tokens, y=avg_cumulative, col sep=comma] {figures/cad_token_cumulative_coverage.csv};
				\addplot[
				dashed,
				color=gray,
				domain=0:100
				] {x/100};
				\addlegendentry{Observed}
				\addlegendentry{Uniform}
			\end{axis}
		\end{tikzpicture}
		\caption{\textbf{Token Utilization}}
		\label{fig:token_analysis_b}
	\end{subfigure}
	\hfill
	\begin{subfigure}{0.32\linewidth}
		\centering
		\begin{tikzpicture}
			\begin{axis}[
				width=1.15\linewidth,
				height=\linewidth,
				xlabel={Prompts $P$},
				ylabel={Time/image [s]},
				xmin=0, xmax=36,
				ymin=0, ymax=0.5,
				grid=both,
				grid style={line width=0.1pt, draw=gray!20},
				major grid style={line width=0.2pt, draw=gray!40},
				tick label style={font=\tiny},
				label style={font=\tiny},
				]
				\addplot[
				thick,
				smooth,
				mark=none,
				color=eccvblue,
				] coordinates {
					(2,0.047)
					(3,0.060)
					(8,0.117)
					(21,0.269)
					(28,0.353)
					(30,0.375)
					(33,0.410)
				};
				\addplot[
				only marks,
				mark=*,
				mark size=1.2pt,
				color=eccvblue,
				] coordinates {
					(2,0.047)
					(3,0.060)
					(8,0.117)
					(21,0.269)
					(28,0.353)
					(30,0.375)
					(33,0.410)
				};
			\end{axis}
		\end{tikzpicture}
		\caption{\textbf{Inference Scaling}}
		\label{fig:token_analysis_c}
	\end{subfigure}
	\caption{\textbf{Latent token analysis and inference scaling.} (a)~Gini coefficient of the ICMT Stage~1 cross-attention from latent tokens $\to$ query image patches increases across ICMT layers, showing progressive spatial grounding onto the prompted object. (b)~Cumulative detection distribution across tokens shows near-uniform utilization rather than mode collapse. (a,\,b) computed on the Clutter6D val.\ set (50k images, 175k matched predictions). (c)~Per-image inference time (H200) across the seven BOP-Classic-Core datasets (dots) scales approximately linearly with object prompts~$P$.}
	\label{fig:token_analysis}
\end{figure}

We evaluate the Gini coefficient (attention concentration) of the Stage~1 cross-attention weights, where latent tokens attend to query image patches, across all three ICMT layers.
As visible in \cref{fig:token_analysis_a}, there is a monotonic increase, indicating that the attention distribution shifts from relatively diffuse to highly concentrated.
This progressive sharpening reflects iterative grounding: early layers gather information broadly across many spatial locations of the query image, while later layers narrow focus toward patches corresponding to the prompted object.

\subsection{Computation Discussion on Naive ICMT Variant}

As highlighted in~Sec.~3.2, employing learnable latent tokens drastically reduces computational overhead in the ICMT.
Given $H_{img}{=}W_{img}{=}720$ and \linebreak $H_{tmpl}{=}W_{tmpl}{=}224$, we obtain $N_{img}{=}2,025$ query image patches and \linebreak $N_{tmpl}{=}N_v\times196{=}2,352$ template patches.
Direct cross-attention results in \linebreak $\mathcal{O}(N_{img}~\times~N_{tmpl})\approx4.8$ million attention computations per layer.
Employing $N_L{=}100$ learnable latent tokens as a compression bottleneck, which attend to both template and image features in an interleaving fashion, instead reduces this computation to $\mathcal{O}(N_L \times (N_{tmpl}+N_{img}))\approx0.4$ million operations.
The cross-attention of image features to latent tokens additionally adds $\mathcal{O}(N_{img}\times N_L)\approx0.2$ million ops, resulting in approximately $0.6$ million operations in total~---~a reduction of around 8 times.

\section{Clutter6D}
\label{sec:clutter6d}

\subsection{Rendering Settings}

\Cref{tab:dataset_params} summarizes all parameters used for generating the Clutter6D dataset, including scene composition, object placement, camera sampling, and material/lighting settings. \Cref{fig:clutter6d_samples_appendix} provides additional sample images from both dataset subsets.

\begin{table}[ht]
	\caption{\textbf{Summary of the dataset generation parameters and sampling distributions.} Parameters shown as HLD / IND indicate values that differ between the two dataset subsets.}
	\centering
	\tiny
	\setlength{\tabcolsep}{4pt}
	\begin{tabularx}{\linewidth}{@{}l l l X@{}l}
		\toprule
		\textbf{Parameter} & \textbf{Value / Range} & \textbf{Distribution} & \textbf{Notes} \\
		\midrule
		\multicolumn{4}{@{}l}{\textit{Scene parameters}} \\
		Images per scene & 25 & fixed & \\
		Resolution & \(720\times720\) & fixed & Divisible by 16 (ViT compatibility) \\
		Number of background textures & 1{,}219 & \(\mathcal{U}\{1,1219\}\) & One sampled per scene \\
		\midrule
		\multicolumn{4}{@{}l}{\textit{Object parameters (Clutter6D-HLD / Clutter6D-IND)}} \\
		Classes per scene & 10--20 / 2--5 & \(\mathcal{U}\) & Distinct object classes \\
		Max instances per class & 10 / 50 & hard cap & Per-class instance limit \\
		Max instances per scene & 40 / 60 & hard cap & Global instance limit \\
		Instances per class & \(0\)--\(10\) / \(0\)--\(50\) & TPois & HLD:\(\text{Pois}(3.5){-}2\); IND:\(\text{Pois}(40){-}20\) \\
		Class size [m] & \([0.1,\,0.3]\) & \(\mathcal{U}[0.1,0.3]\) & Same within class (per scene) \\
		\midrule
		\multicolumn{4}{@{}l}{\textit{Position sampling}} \\
		Radial span [m] & \([0.2,\,0.5]\) & \(\mathcal{U}[0.2,0.5]\) & Initial XY sampling shell \\
		Drop height [m] & \([0.4,\,0.6]\) & \(\mathcal{U}[0.4,0.6]\) & Rigid-body settle to induce clutter \\
		\midrule
		\multicolumn{4}{@{}l}{\textit{Camera parameters}} \\
		Camera radius [m] & \([0.5,\,1.8]\) & \(\mathcal{U}[0.5,1.8]\) & Distance from scene center \\
		Camera elevation [\(^\circ\)] & \([5,\,89]\) & \(\mathcal{U}[5,89]\) & Azimuth uniform on viewing shell \\
		\(f_x\) [px] & \([400,\,2000)\) & \(\mathcal{U}[400,2000)\) & Focal length in pixels \\
		\(f_y\) [px] & \(f_x{+}\Delta\) & \(\Delta\!\sim\!\mathcal{U}[-3,3]\) & Mild anisotropy \\
		Principal point [px] & \((W/2,\ H/2)\) & fixed & Image center \\
		\midrule
		\multicolumn{4}{@{}l}{\textit{Material and lighting}} \\
		Roughness & \([0.0,\,0.9]\) & \(\mathcal{U}[0.0,0.9]\) & Per instance \\
		Metallic & \([0.0,\,0.9]\) & \(\mathcal{U}[0.0,0.9]\) & Per instance \\
		Light color (RGB) & \([0.5,\,1.0]^3\) & \(\mathcal{U}([0.5,1]^3)\) & Per scene \\
		Light pose & \(r\!\in\![1.0,1.5],\ e\!\in\![5,89]^\circ\) & uniform on shell & Around scene center \\
		\bottomrule
	\end{tabularx}
	\label{tab:dataset_params}
\end{table}

\subsection{Objects, Classes and their Distributions}

\begin{figure}[ht!]
	\centering
	\begin{subfigure}[t]{0.36\textwidth}
		\centering
		\includegraphics[width=\textwidth]{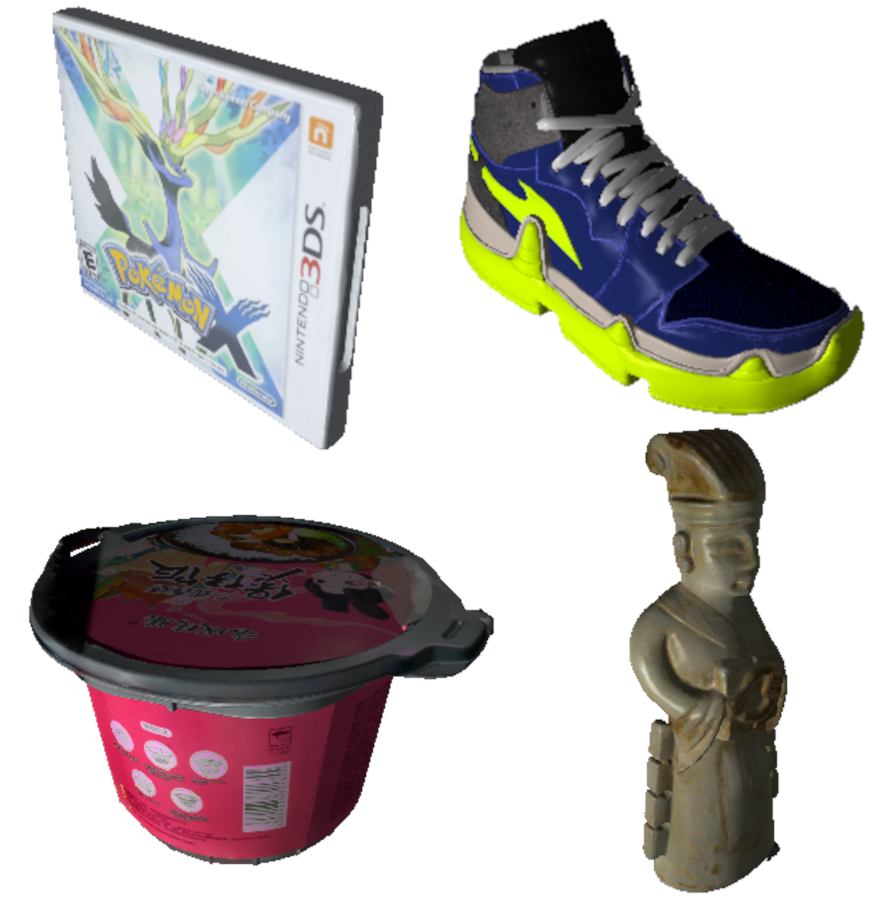}
		\caption{3D model samples from the Clutter6D object catalog.}
		\label{fig:models}
	\end{subfigure}
	\hfill
	\begin{subfigure}[t]{0.62\textwidth}
		\centering
		\includegraphics[width=\textwidth]{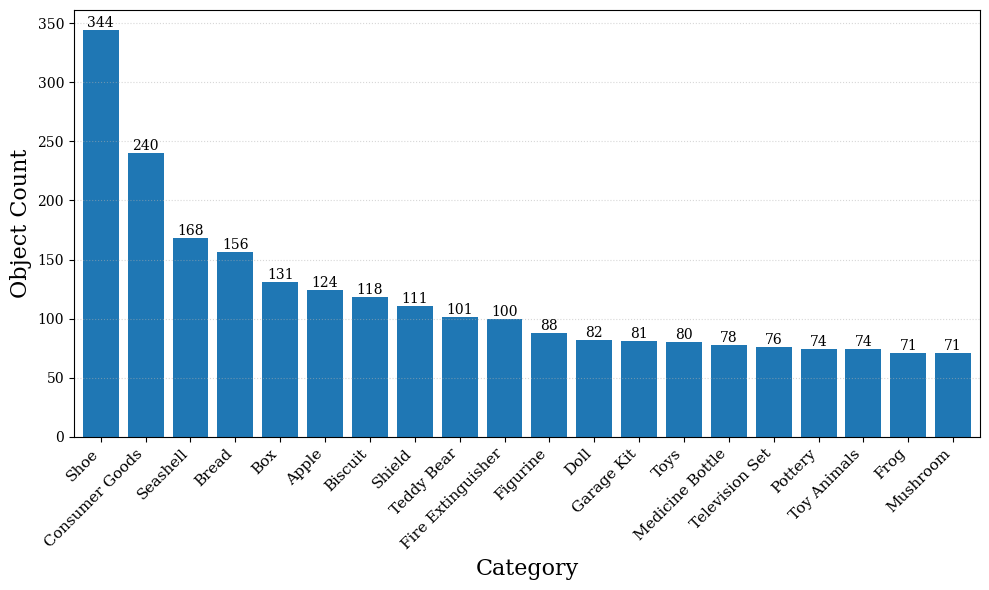}
		\caption{Top 20 most frequent object categories in the object catalog of Clutter6D.}
		\label{fig:categories}
	\end{subfigure}
	\caption{\textbf{Clutter6D object catalog overview.} (a) Samples from the Clutter6D object catalog, top-left from gso (category: Media Cases), top-right from Objaverse (category: Shoe), bottom-left from oo3d (category: Bucket Noodle), bottom-right from oo3d (category: Figurine). (b) Top 20 categories by frequency in the catalog, selected from 681 total categories with a mean of 14.2 objects per category, unlabelled objects excluded.}
	\label{fig:clutter6d_cats_and_models}
\end{figure}

Example object models are shown in~\cref{fig:models}, and \cref{fig:categories} lists the 20 most frequent categories across the whole dataset.

\begin{figure}[t]
	\centering
	\begin{subfigure}[t]{0.315\textwidth}
		\centering
		\includegraphics[width=0.94\textwidth]{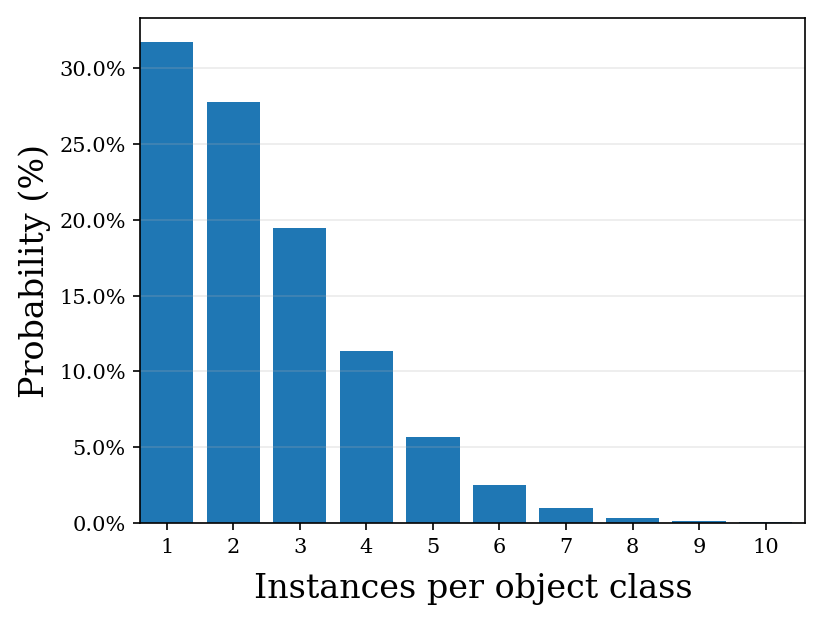}
		\caption{Clutter6D-HLD: Per-class instance Poisson distribution.}
		\label{fig:poisson_hld}
	\end{subfigure}
	\hfill
	\begin{subfigure}[t]{0.675\textwidth}
		\centering
		\includegraphics[width=\textwidth]{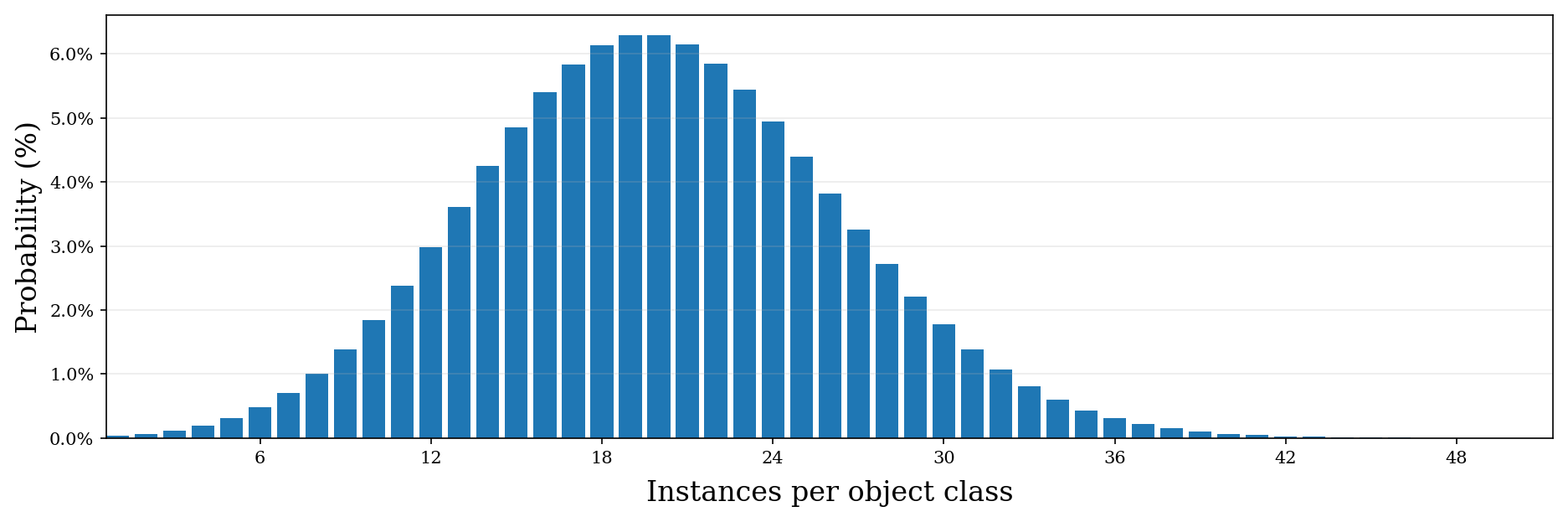}
		\caption{Clutter6D-IND: Per-class instance Poisson distribution.}
		\label{fig:poisson_ind}
	\end{subfigure}
	\caption{\textbf{Object instance sampling distributions.} Probability mass functions controlling the number of object instances per class in synthetic scenes for both dataset variants.}
	\label{fig:poisson}
\end{figure}

\begin{figure}[htb!]
	\centering
	\begin{subfigure}[t]{\textwidth}
		\centering
		\includegraphics[width=\linewidth]{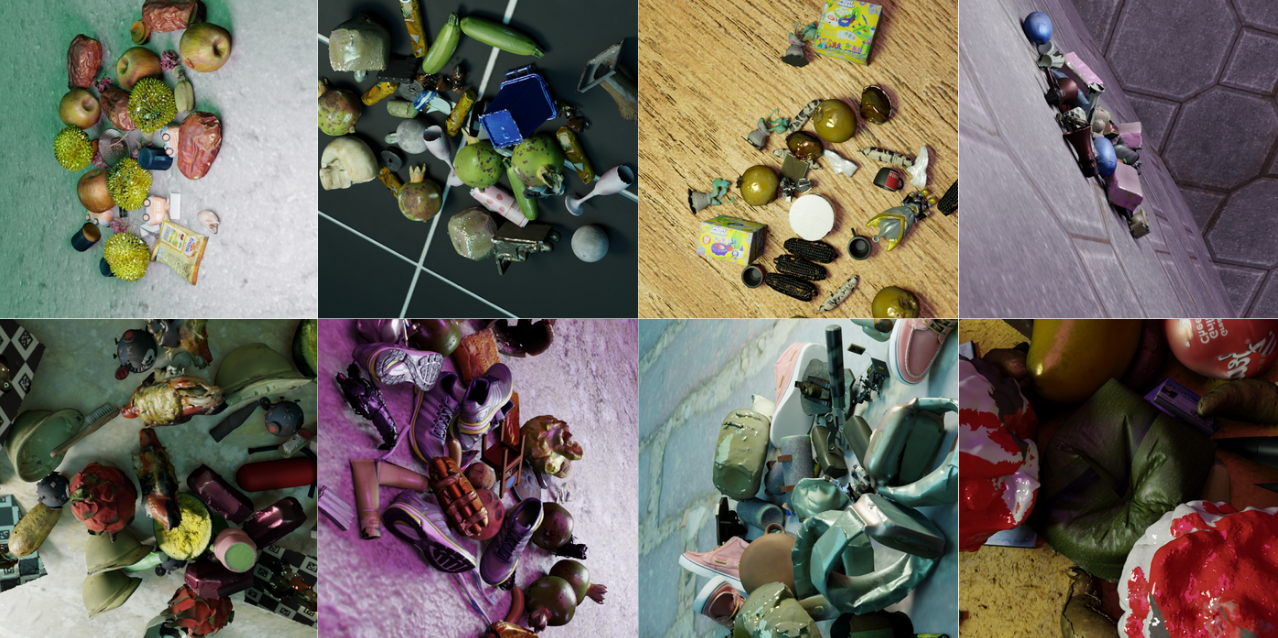}
		\caption{Clutter6D-HLD samples featuring diverse object classes with fewer instances per class.}
	\end{subfigure}
	
	\vspace{0.5em}
	
	\begin{subfigure}[t]{\textwidth}
		\centering
		\includegraphics[width=\linewidth]{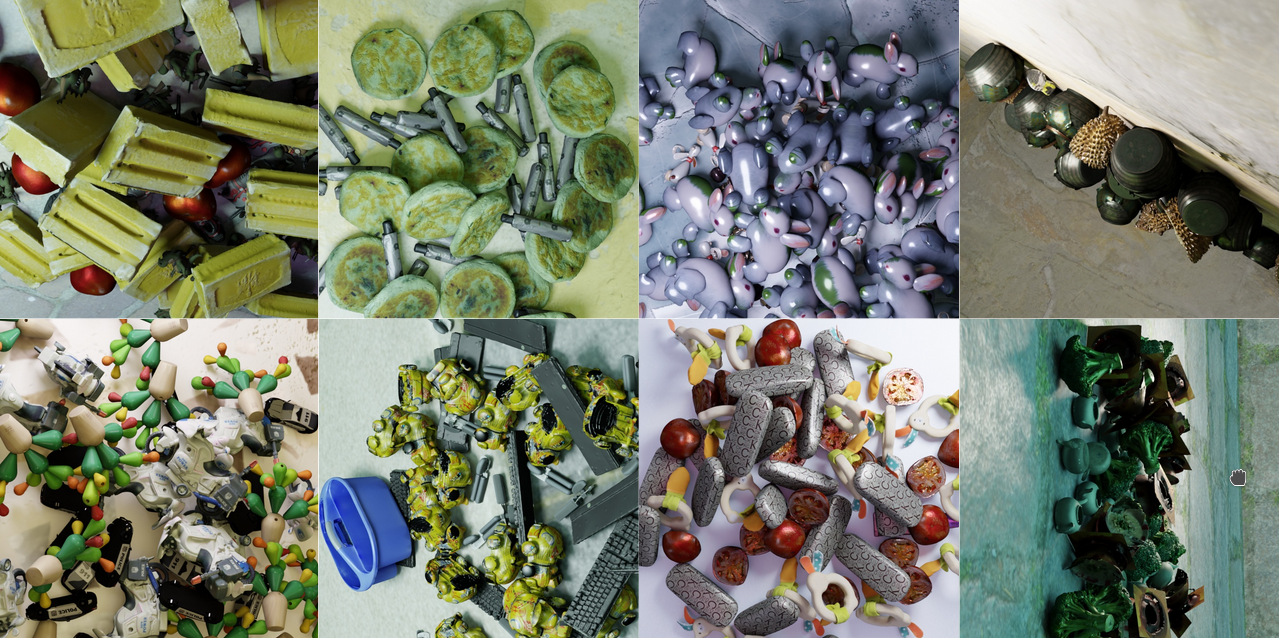}
		\caption{Clutter6D-IND samples featuring fewer object classes with more instances per class.}
	\end{subfigure}
	\caption{\textbf{Additional Clutter6D query image samples.}}
	\label{fig:clutter6d_samples_appendix}
\end{figure}

The two subsets of Clutter6D, the household and the industrial splits, differ only in the number of distinct object classes per scene and in the number of instances per class.
The household subset recreates typical household environments where many different object classes appear with fewer multiple instances. Between 10 and 20 object classes are uniformly sampled, and the scene is populated with instances whose per-class counts are randomly sampled from a truncated and shifted Poisson distribution (see~\cref{fig:poisson_hld}). 
Counts are capped at 10 instances per class and at 40 instances per scene in total.
The industrial subset recreates typical industrial environments where fewer object classes appear with more multiple instances, as in bin-picking tasks. 
Between 2 and 5 object classes are uniformly sampled, and per-class instance counts are drawn from a truncated, shifted Poisson distribution with different parameters than in Clutter6D-HLD (see~\cref{fig:poisson_ind}). 
Counts are capped at 50 instances per class and at 60 instances per scene in total.

\subsection{Template Rendering}

For CAD template rendering during onboarding (cf.\ Sec.~3.1 of the main paper), we employ a lightweight PyTorch3D-based~\cite{ravi2020pytorch3d} renderer rather than BlenderProc's physically-based rendering pipeline. Each CAD model is rendered into 42 views from viewpoints sampled via a Fibonacci sphere, providing uniform coverage of the object's geometry. We pre-render 42 views to facilitate ablations over different template counts. For our final model, we select $N_v{=}12$ views via farthest-point sampling (see \cref{tab:hyperparams}), meaning that at inference time only 12 templates need to be rendered per object, well under 1\,s on an NVIDIA H200 GPU. The background is set to the ImageNet~\cite{deng2009imagenet} channel-wise mean, so that background pixels become zero after the backbone's ImageNet normalization, thereby preventing the network from learning spurious correlations with artificial background boundaries. This rapid rendering enables fast onboarding of new objects, a practical advantage in settings where object inventories change frequently. Each object's templates are stored as a self-contained HDF5 file, allowing selective on-demand loading of only those templates that appear in a given query image.

\newpage
\section{Miscellaneous}
\label{sec:misc}

\subsection{ICMT Features on T-LESS}

\begin{figure}[ht]
	\centering
	\includegraphics[width=.6\linewidth]{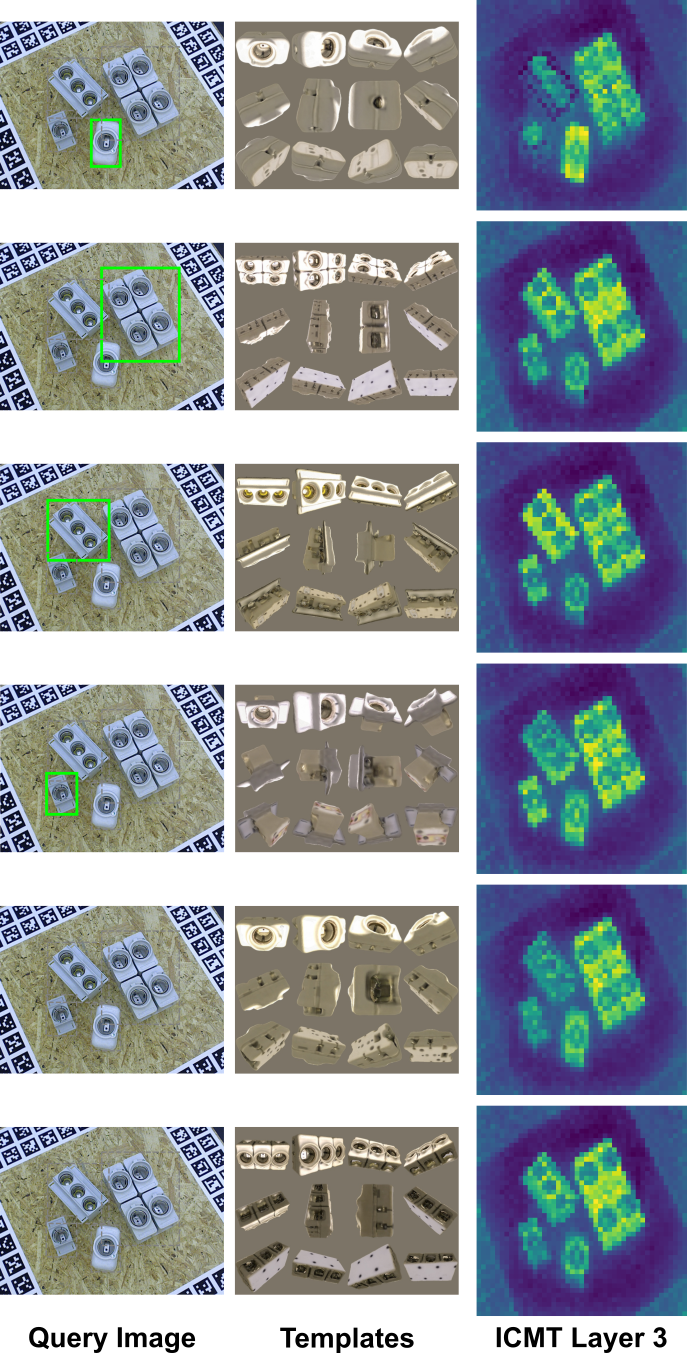}
	\caption{\textbf{ICMT features on T-LESS.} The first four rows show positive object prompts, while the last two rows show prompts of objects that are not in the query image. Note that the ICMT features in layer 3 are highly similar across rows.
	}
	\label{fig:tless}
\end{figure}

Compared to Fig. 5 in the main paper, the ICMT finds it more difficult to clearly separate the similar object appearances of T-LESS.
We visualize an example scene in~\cref{fig:tless}.

\subsection{Details on the CAD Model Degradation}

\begin{figure}[ht]
	\centering
	\includegraphics[width=.8\linewidth]{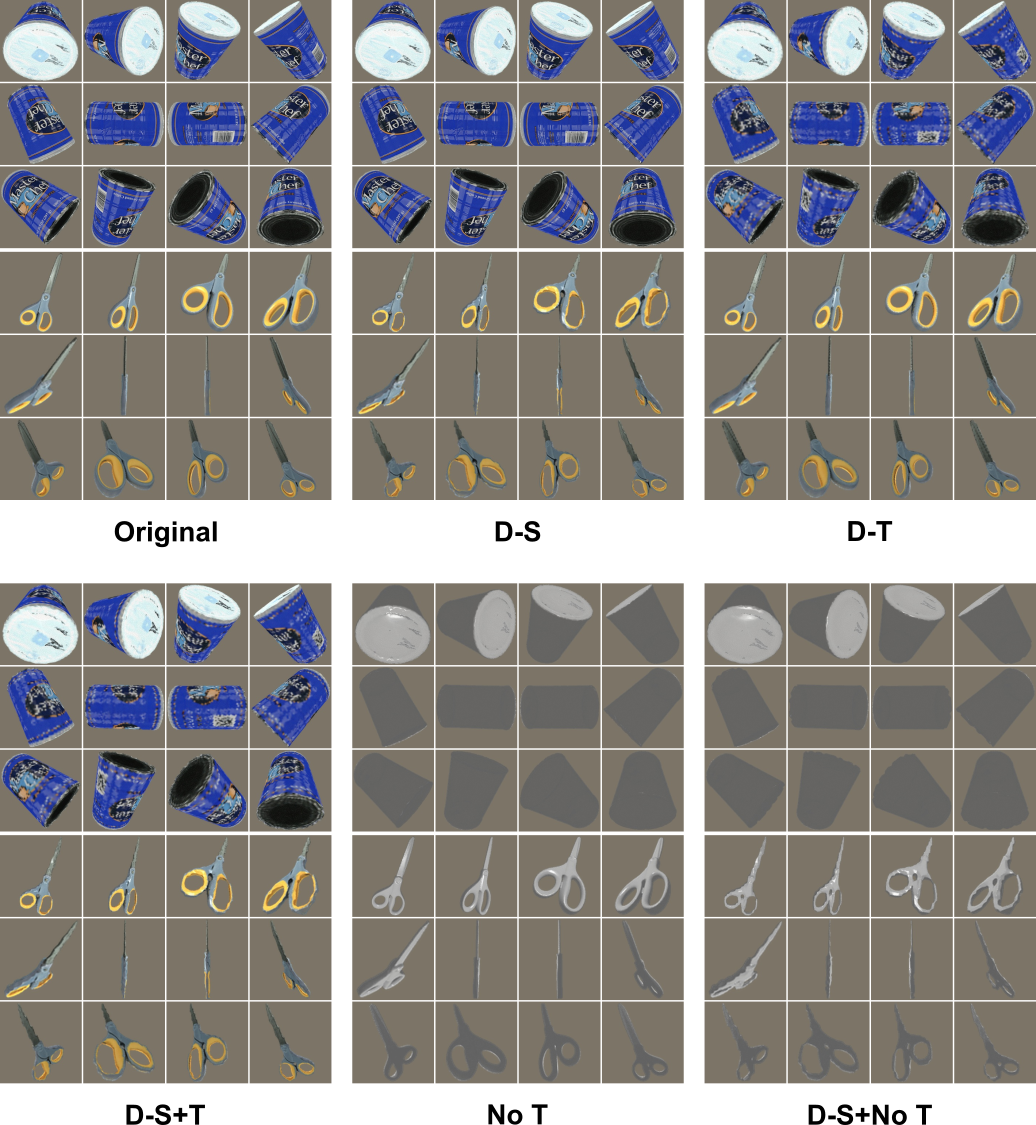}
	\caption{\textbf{CAD model degradation.} Degraded shape and texture examples for \emph{master\_chef\_can} and \emph{scissors}.
	}
	\label{fig:cad_deg}
\end{figure}

Sec. 5.3 investigates the performance of FOMO on degraded CAD models, both in terms of texture and mesh geometry.
For the former, we resize the 4k textures to a resolution of $128\times128$ pixels using cubic interpolation.
Regarding the latter, we apply a remeshing modifier on an object mesh's clone (\emph{mode}=\emph{"smooth"}, \emph{octree\_depth}=\emph{5}, \emph{scale}=\emph{0.9} and \emph{use\_remove\_disconnected}=\emph{False}) followed by a shrinkwrap modifier (default settings) on the original mesh with the clone as target to keep valid uv texture mappings.
Operations are performed with BlenderProc~\cite{denninger2023blenderproc2} and Blender 4.2~\cite{blender_4_2_release}.
\Cref{tab:cad_model_deg_full} lists the detailed results of the CAD model degradation study, and~\cref{fig:cad_deg} shows qualitative examples of the PyTorch3D rendered templates.

\begin{table}[ht]
	\centering
	\caption{Per-class AP values. 
		D-S = degraded shape; 
		D-T = degraded texture.}
	\scriptsize

	\begin{tabularx}{\textwidth}{c *{7}{>{\centering\arraybackslash}X}}
		\toprule
		Object & Type & Baseline & D-S & D-T & D-S+T & No T & D-S+No T \\
		\midrule
		2 & box & 82.4 & 84.5 & 70.4 & 78.4 & 0.0 & 0.0 \\
		3 & box & 84.4 & 82.8 & 84.7 & 82.6 & 0.0 & 0.0 \\
		7 & box & 58.1 & 60.9 & 38.4 & 33.9 & 0.0 & 0.0 \\
		8 & box & 77.6 & 82.3 & 16.5 & 20.4 & 0.0 & 0.0 \\
		9 & box & 62.3 & 62.5 & 60.2 & 57.0 & 0.7 & 0.2 \\
		16 & box & 71.0 & 71.1 & 71.0 & 71.3 & 6.6 & 21.8 \\
		21 & box & 81.1 & 80.6 & 80.6 & 79.7 & 78.5 & 79.1 \\
		\midrule
		Box Mean &  & 73.8 & 75.0 & 60.3 & 60.5 & 12.3 & 14.4 \\
		\midrule
		1 & cylindric & 75.2 & 72.4 & 12.1 & 2.6 & 0.4 & 0.3 \\
		4 & cylindric & 76.3 & 76.4 & 66.0 & 73.6 & 0.3 & 0.0 \\
		6 & cylindric & 81.6 & 81.9 & 81.8 & 81.7 & 16.4 & 5.4 \\
		18 & cylindric & 63.2 & 69.6 & 54.4 & 63.3 & 33.3 & 13.0 \\
		\midrule
		Cyl. Mean &  & 74.1 & 75.1 & 53.6 & 55.3 & 12.6 & 4.7 \\
		\midrule
		5 & misc & 88.6 & 88.4 & 88.5 & 88.4 & 87.8 & 83.3 \\
		10 & misc & 55.9 & 56.6 & 56.5 & 59.9 & 63.4 & 66.1 \\
		11 & misc & 89.8 & 88.3 & 89.6 & 87.5 & 61.1 & 21.0 \\
		12 & misc & 80.2 & 80.3 & 78.4 & 79.4 & 66.9 & 49.3 \\
		13 & misc & 84.6 & 84.2 & 84.9 & 83.3 & 84.3 & 38.9 \\
		14 & misc & 75.2 & 61.1 & 75.0 & 64.2 & 77.5 & 51.7 \\
		15 & misc & 84.9 & 83.9 & 83.7 & 81.8 & 33.9 & 6.2 \\
		17 & misc & 36.9 & 64.8 & 34.2 & 60.1 & 46.7 & 54.4 \\
		19 & misc & 44.6 & 29.9 & 33.2 & 25.0 & 43.4 & 4.1 \\
		20 & misc & 43.1 & 24.4 & 41.2 & 21.7 & 31.2 & 0.9 \\
		\midrule
		Misc Mean &  & 68.4 & 66.2 & 66.5 & 65.1 & 59.6 & 37.6 \\
		\midrule
		\textbf{Total Mean} &  & 71.2 & 70.8 & 62.0 & 61.7 & 34.9 & 23.6 \\

		\bottomrule
	\end{tabularx}
	\label{tab:cad_model_deg_full}
\end{table}

\FloatBarrier


\end{document}